%% file: 0-main.tex
\documentclass[10pt,twocolumn,letterpaper]{article}

\usepackage{cvpr}              

\usepackage{graphicx}
\usepackage{amsmath}
\usepackage{amssymb}
\usepackage{booktabs}
\usepackage[usenames,dvipsnames]{color}

\usepackage{circledsteps}

\DeclareMathOperator*{\argmin}{arg\,min}

\usepackage{epsfig}
\usepackage{graphicx}
\usepackage{amsmath}
\usepackage{amssymb}
\usepackage{mathtools}
\usepackage{subcaption}
\usepackage{comment}
\usepackage{mathtools}
\usepackage{acronym}
\usepackage{caption}
\usepackage{array}
\usepackage{tabularx}
\usepackage{subcaption}
\usepackage{booktabs}
\usepackage{multicol}
\usepackage{multirow}
\usepackage[norelsize, linesnumbered, ruled, lined, boxed, commentsnumbered]{algorithm2e}

\usepackage[pagebackref=true,breaklinks=true,letterpaper=true,colorlinks,bookmarks=false]{hyperref}

\usepackage{xspace}

\newcommand{\red}[1]{\textcolor{black}{#1}}

\usepackage[capitalize]{cleveref}
\crefname{section}{Sec.}{Secs.}
\Crefname{section}{Section}{Sections}
\Crefname{table}{Table}{Tables}
\crefname{table}{Tab.}{Tabs.}

\def\cvprPaperID{1082} 
\def\confName{CVPR}
\def\confYear{2023}

\begin{document}
\vspace{-4mm}


\title{3DAvatarGAN: Bridging Domains for Personalized Editable Avatars}

\author{Rameen Abdal\textsuperscript{†1} \quad Hsin-Ying Lee\textsuperscript{2} \quad Peihao Zhu\textsuperscript{†1}  \quad Menglei Chai\textsuperscript{2} \quad Aliaksandr Siarohin\textsuperscript{2} 
\\ \\
 \quad Peter Wonka\textsuperscript{1} \quad Sergey Tulyakov\textsuperscript{2} \\
\\
\textsuperscript{1}KAUST \quad  \textsuperscript{2}Snap Inc.
}

\twocolumn[{%
\renewcommand\twocolumn[1][]{#1}%
\maketitle
\thispagestyle{empty}
\vspace{-10mm}
\begin{center}
     \includegraphics[width= 0.99\linewidth]{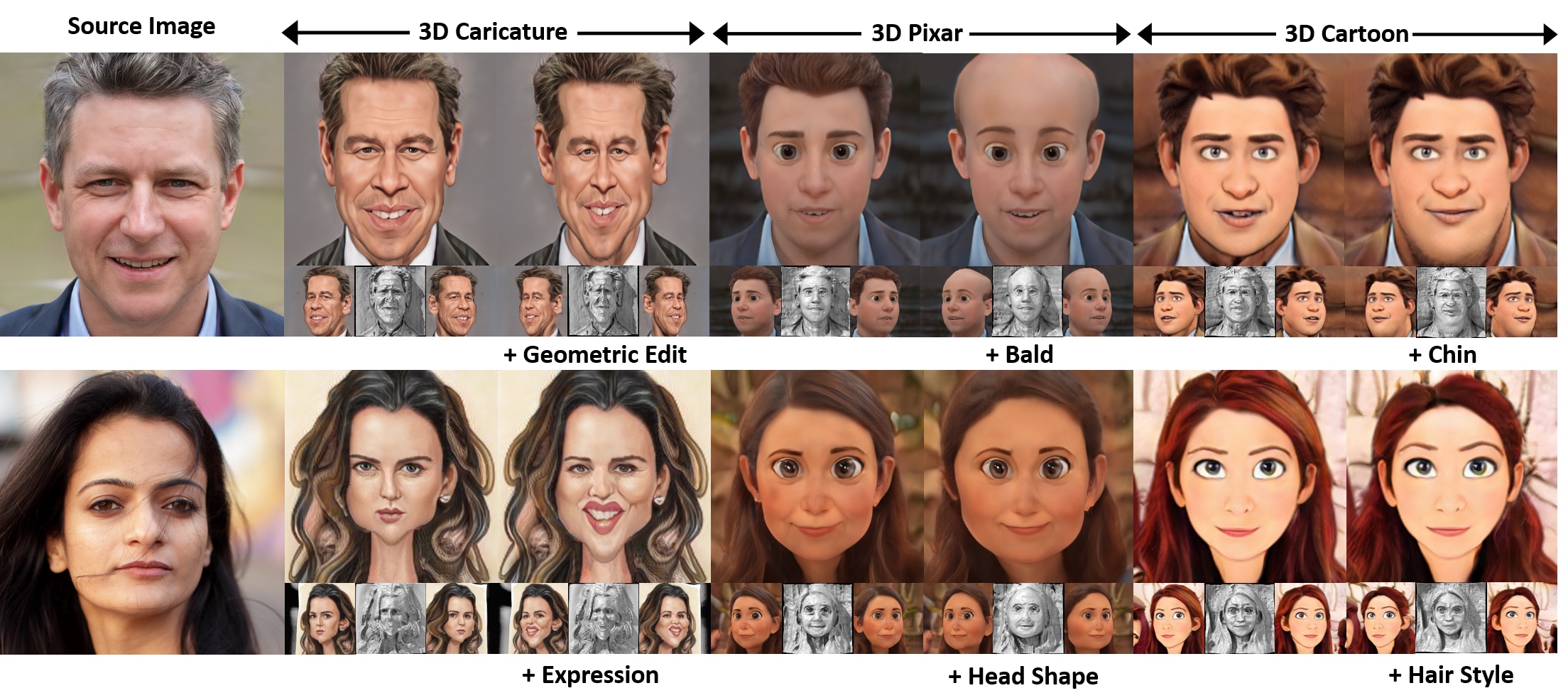}
     \captionof{figure}{ \textbf{Editable 3D avatars. } We present 3DAvatarGAN, a 3D GAN able to produce and edit personalized 3D avatars from a single photograph (real or generated). Our method distills information from a 2D-GAN trained on 2D artistic datasets like Caricatures, Pixar toons, Cartoons, Comics \etc and requires no camera annotations.
     }
    \label{fig:teaser}
\end{center}
}]
\begin{abstract}
Modern 3D-GANs synthesize geometry and texture by training on large-scale datasets with a consistent structure. Training such models on stylized, artistic data, 
with often unknown, highly variable geometry, and camera information has not yet been shown possible.
Can we train a 3D GAN on such artistic data, while maintaining multi-view consistency and texture quality?
To this end, we propose an adaptation framework, where the source domain is a pre-trained 3D-GAN, while the target domain is a 2D-GAN trained on artistic datasets. We, then, distill the knowledge from a 2D generator to the source 3D generator. To do that, we first propose an optimization-based method to align the distributions of camera parameters across domains. Second, we propose regularizations necessary to learn high-quality texture, while avoiding degenerate geometric solutions, such as flat shapes. Third, we show a deformation-based technique for modeling exaggerated geometry of artistic domains, enabling---as a byproduct---personalized geometric editing. Finally, we propose a novel inversion method for 3D-GANs linking the latent spaces of the source and the target domains. Our contributions---for the first time---allow for the generation, editing, and animation of personalized artistic 3D avatars on artistic datasets.
Project Page:  \href{https:/rameenabdal.github.io/3DAvatarGAN}{https:/rameenabdal.github.io/3DAvatarGAN}

\end{abstract}

\input{1-intro-2}

\input{2-related}
\input{3-method}

\input{4-experiment}
\input{5-conclusion}
\clearpage
\input{suppl}


\clearpage
{\small
\bibliographystyle{ieee_fullname}
\bibliography{egbib}
}

\end{document}

%% file: 1-intro-2.tex
\vspace*{-0.5cm}

\section{Introduction}
\label{sec:intro}
\vspace{-5mm}
\footnote[0]{\textsuperscript{†} Part of the work was done during an internship at Snap Inc.}

Photo-realistic portrait face generation is an iconic application demonstrating the capability of generative models especially GANs~\cite{karras2019style, Karras2020ada, karras2021aliasfree}. A recent development has witnessed an advancement from straightforwardly synthesizing 2D images to learning 3D structures without 3D supervision, referred to as 3D-GANs~\cite{EG3D,  VolumeGAN, StyleSDF, ide-3d}.
Such training is feasible with the datasets containing objects with highly consistent geometry, enabling a 3D-GAN to learn a distribution of shapes and textures. In contrast, artistically stylized datasets~\cite{Jang2021StyleCari,yang2022Pastiche } have arbitrary exaggerations of both geometry and texture, for example, the nose, cheeks, and eyes can be arbitrarily drawn, depending on the style of the artist as well as on the features of the subject, see Fig.~\ref{fig:teaser}. Training a 3D-GAN on such data becomes problematic due to the challenge of learning such an arbitrary distribution of geometry and texture. In our experiments (Sec.~\ref{sec:quant}), 3D-GANs~\cite{EG3D} generate flat geometry and become 2D-GANs essentially. A natural question arises, whether a 3D-GAN can synthesize consistent novel views of images belonging to artistically stylized domains, such as the ones in Fig.~\ref{fig:teaser}.

In this work, we propose a domain-adaption framework that allows us to answer the question positively. Specifically, we fine-tune a pre-trained 3D-GAN using a 2D-GAN trained on a target domain. Despite being well explored for 2D-GANs~\cite{Jang2021StyleCari,yang2022Pastiche}, existing domain adaptation techniques are not directly applicable to 3D-GANs, due to the nature of 3D data and characteristics of 3D generators. 

The geometry and texture of stylized 2D datasets can be arbitrarily exaggerated depending on the context, artist, and production requirements. Due to this, no reliable way to estimate camera parameters for each image exists, whether using an off-the-shelf pose detector~\cite{githubrepo} or a manual labeling effort. To enable the training of 3D-GANs on such challenging datasets, we propose three contributions. \Circled{\textbf{1}} An optimization-based method to align distributions of camera parameters between domains. \Circled{\textbf{2}} Texture, depth, and geometry regularizations to avoid degenerate, flat solutions and ensure high visual quality. Furthermore, we redesign the discriminator training to make it compatible with our task. We then propose \Circled{\textbf{3}} a \textit{Thin Plate Spline (TPS)} 3D deformation module operating on a tri-plane representation to allow for certain large and sometimes extreme geometric deformations, which are so typical in artistic domains.

The proposed adaptation framework enables the training of 3D-GANs on complex and challenging artistic data. The previous success of domain adaptation in 2D-GANs unleashed a number of exciting applications in the content creation area~\cite{Jang2021StyleCari,yang2022Pastiche}. Given a single image such methods first find a latent code corresponding to it using GAN inversion, followed by latent editing producing the desired effect in the image space. Compared to 2D-GANs, the latent space of 3D-GANs is more entangled, making it more challenging to link the latent spaces between domains, rendering the existing inversion and editing techniques not directly applicable.
Hence, we take a step further and explore the use of our approach to 3D artistic avatar generation and editing. Our final contribution to enable such applications is \Circled{\textbf{4}} a new inversion method for coupled 3D-GANs.

In summary, the proposed domain-adaption framework allows us to train 3D-GANs on challenging artistic datasets with exaggerated geometry and texture. We call our method 3DAvatarGAN as it---for the first time---offers generation, editing, and animation of personalized stylized, artistic avatars obtained from a single image. Our results (See Sec.~\ref{sec:qual}) show the high-quality 3D avatars possible by our method compared to the naive fine-tuning.


%% file: 2-related.tex
\section{Related Work}


\noindent\textbf{GANs and Semantic Image Editing.} Generative adversarial Networks (GANs)~\cite{goodfellow2014generative,radford2015unsupervised} are one popular type of generative model, especially for smaller high-quality datasets such as FFHQ~\cite{STYLEGAN2018}, AFHQ~\cite{choi2020starganv2}, and LSUN objects~\cite{yu15lsun}. For these datasets, StyleGAN~\cite{STYLEGAN2018,karras2021aliasfree,Karras2020ada} can be considered as the current state-of-the-art GAN~\cite{karras2017progressive,STYLEGAN2018,Karras2019stylegan2,Karras2020ada,karras2021aliasfree}. The disentangled latent space learned by StyleGAN has been shown to exhibit semantic properties conducive to semantic image editing~\cite{shen2020interfacegan,abdal2019image2stylegan,patashnik2021styleclip,abdal2020styleflow, harkonen2020ganspace, tewari2020stylerig,wu2020stylespace,lin2021infinitygan,chong2021stylegan}. CLIP~\cite{DBLP:journals/corr/abs-2103-00020} based image editing~\cite{patashnik2021styleclip,gal2021stylegan,cl2sg} and domain transfer~\cite{zhu2022mind,DBLP:journals/corr/abs-2112-11641} are another set of works enabled by StyleGAN.


\noindent\textbf{GAN Inversion.} Algorithms to project existing images into a GAN latent space are a prerequisite for GAN-based image editing.
There are mainly two types of methods to enable such a projection: optimization-based methods~\cite{abdal2019image2stylegan,zhu2020improved, tewari2020pie,cheng2022inout}
and encoder-based methods~\cite{zhu2020domain, richardson2020encoding, tov2021designing, alaluf2021restyle,DBLP:journals/corr/abs-2111-15666}.
On top of both streams of methods, the generator weights can be further modified after obtaining initial inversion results~\cite{roich2021pivotal}.


\noindent\textbf{Learning 3D-GANs with 2D Data.} Previously, some approaches attempt to extract 3D structure from pre-trained 2D-GANs~\cite{LiftedGAN, 3d-from-2d-GANs}. Recently, inspired by Neural Radiance Field (NeRF)~\cite{NeRF, NeRF++, mip-NeRF, Nerfies}, novel GAN architectures have been proposed to combine implicit or explicit 3D representations with neural rendering techniques~\cite{GRAF, piGAN, CAMPARI, Giraffe,chan2021efficient, StyleNeRF, VolumeGAN, StyleSDF, ide-3d,skorokhodov20233d,xu2022discoscene}.
In our work, we build on EG3D~\cite{chan2021efficient} which has current state-of-the-art results for human faces trained on the FFHQ dataset.





\noindent\textbf{Avatars and GANs.} To generate new results in an artistic domain (\eg anime or cartoons), a promising technique is to fine-tune an existing GAN pre-trained on photographs, e.g.~\cite{https://doi.org/10.48550/arxiv.2010.05334,10.1145/3450626.3459771,wang2022cross}. 
Data augmentation and freezing lower layers of the discriminator are useful tools when fine-tuning a 2D-GAN~\cite{mo2020freeze, Karras2020ada}. One branch of methods~\cite{patashnik2021styleclip,zhu2022mind,gal2021stylegannada} investigates domain adaptation if only a few examples or only text descriptions are available. While others focus on matching the distribution of artistic datasets with diverse shapes and styles. Our work also falls in this domain. Among previous efforts, StyleCariGAN~\cite{Jang2021StyleCari} proposes invertible modules in the generator to train and generate caricatures from real images. DualStyleGAN~\cite{yang2022Pastiche} learns two mapping networks in StyleGAN to control the style and structure of the new domain. Some works are trained on 3D data or require heavy labeling/engineering~\cite{Jung_2022,Ye_2021,han2021exemplar} and use 3D morphable models to map 2D images of caricatures to 3D models. However, such models fail to model the hair, teeth, neck, and clothes and suffer in texture quality. In this work, we are the first to tackle the problem of domain adaption of 3D-GANs and to produce fully controllable 3D Avatars. We employ 2D to 3D domain adaptation and distillation and make use of synthetic 2D data from StyleCariGAN~\cite{Jang2021StyleCari} and DualStyleGAN~\cite{yang2022Pastiche}.


%% file: 3-method.tex
\begin{figure}[t!]
    \centering
    \includegraphics[width=\linewidth]{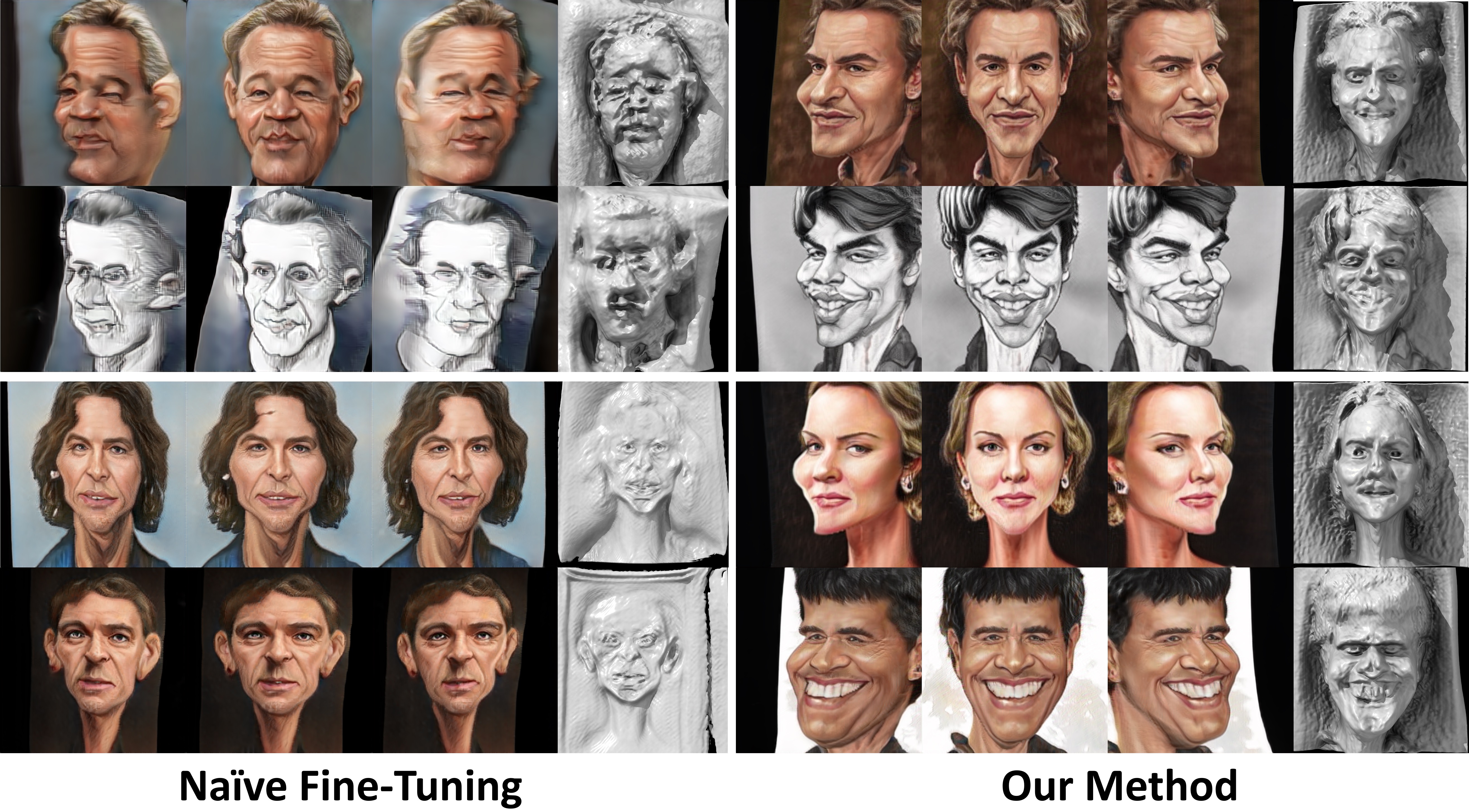}
    \caption{
    \textbf{Comparison with naive fine-tuning.} Comparison of generated 3D avatars with a na\"ivly fine-tuned generator $\mathrm{G_{base}}$ (left sub-figures) versus our generator $\mathrm{G_t}$ (right sub-figures). The corresponding sub-figures show comparisons in terms of texture quality (top two rows) and geometry (bottom two rows). See Sec.~\ref{sec:quant} for details.
   }
     \vspace{-5mm}
    \label{fig:artifacts}
\end{figure} 

\begin{figure}[ht!]

    \centering
    \includegraphics[width=\linewidth]{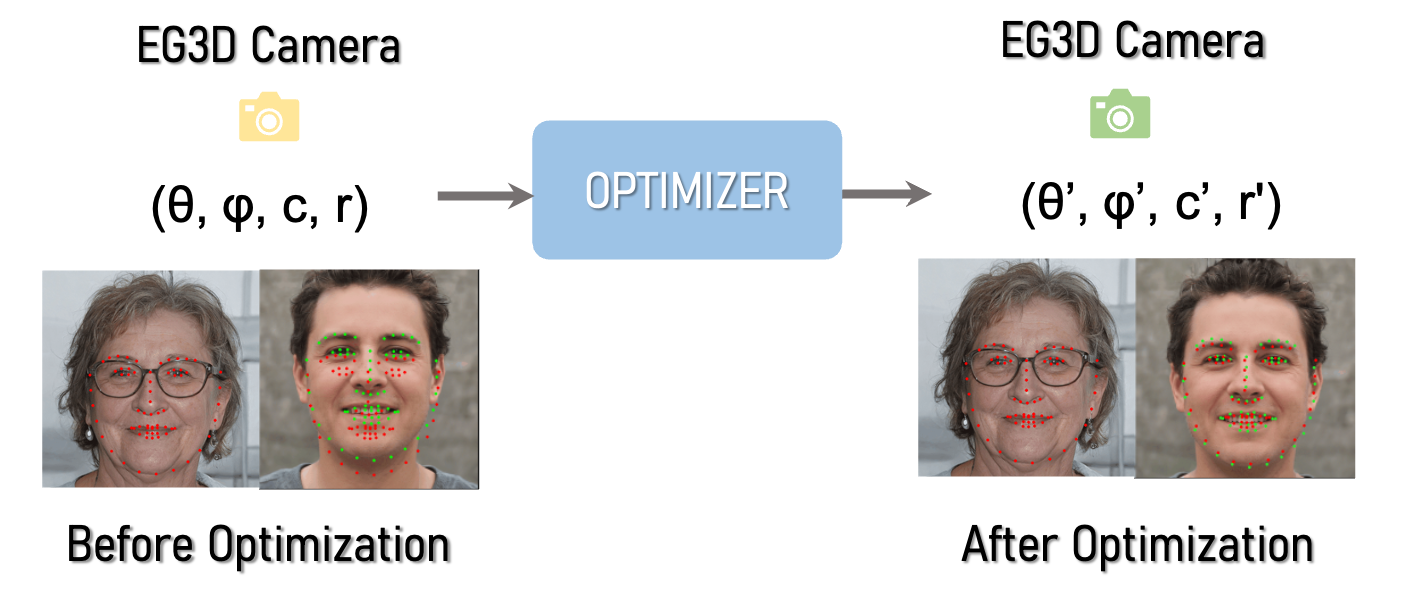}
        \caption{
    \textbf{ Illustration of camera alignment.
   } We define an optimization algorithm to determine the camera parameters of the new domain \ie avatars.}
    \label{fig:camera}
\end{figure}

\begin{figure*}[ht!]

    \centering
    \includegraphics[width=0.85\linewidth]{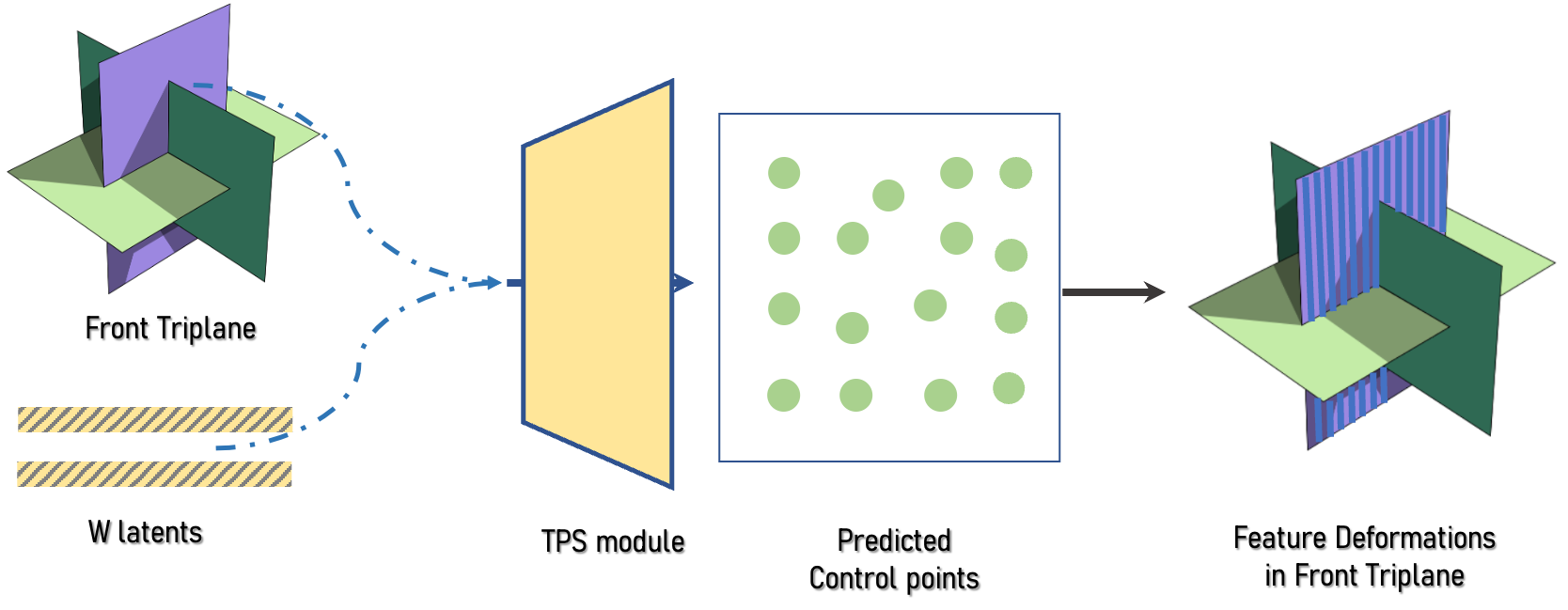}
        \caption{
    \textbf{ Illustration of \textit{TPS} module. 
   } Our \textit{TPS} module takes $W$ latents and the front triplane as inputs and outputs deformed control points that deform the front triplane.}
    \label{fig:tps_inference}
\end{figure*}

\begin{figure*}[ht!]

    \centering
    \includegraphics[width=\linewidth]{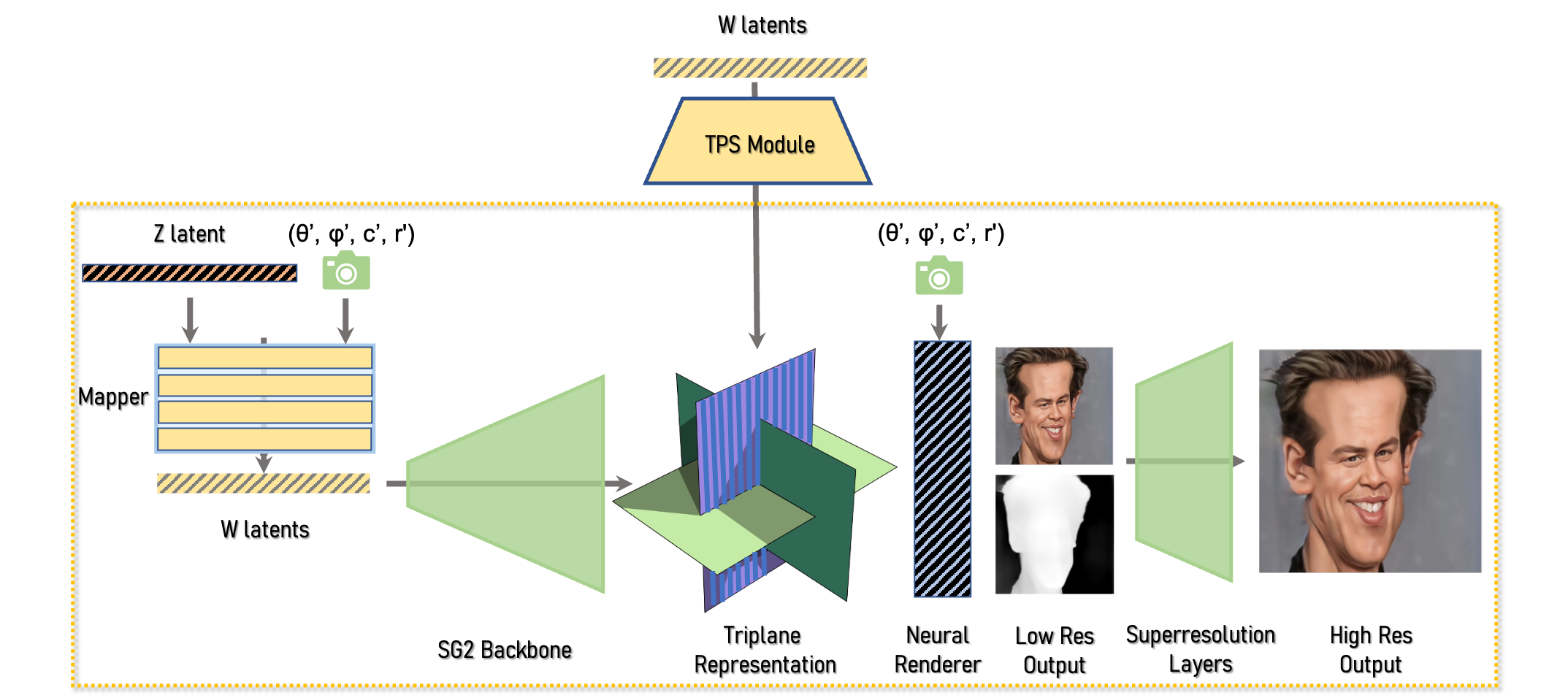}
        \caption{
    \textbf{ Fine tuned EG3D ($\mathrm{G_t}$) pipeline with \textit{TPS} module.
   } We design a framework to fine-tune a 3D-GAN \ie EG3D (dotted yellow box) and adapt to an artistic domain that does not stem from a consistent 3D model. In addition, our framework modifies EG3D architecture  to include a novel \textit{TPS} module for geometric editing. }
    \label{fig:inference}
\end{figure*}

\begin{figure*}[ht!]
    \centering
    \includegraphics[width=\linewidth]{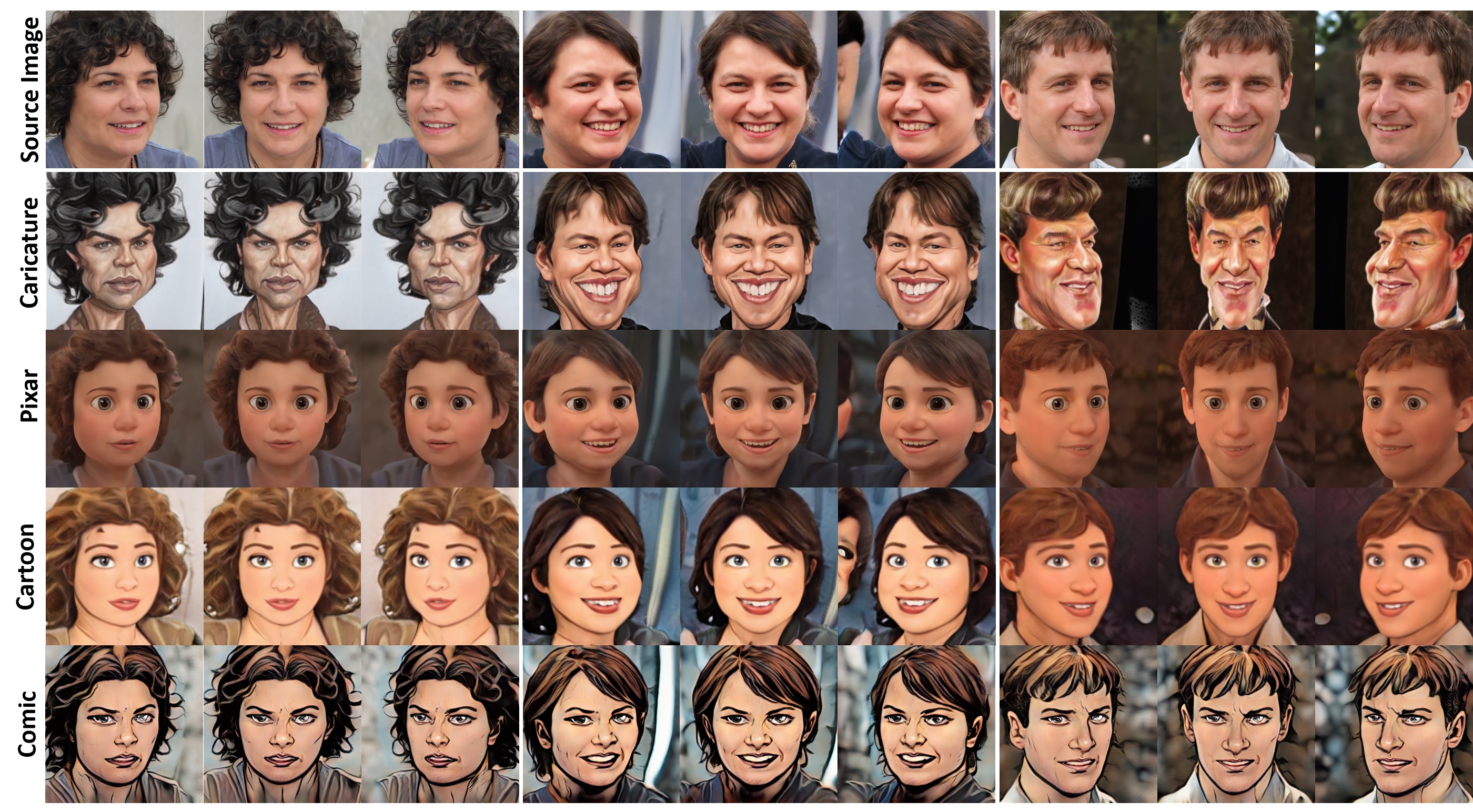}
    \vspace{-5mm}
    \caption{
    \textbf{Domain adaptation.} Domain adaptation results of images from source domain $T_\mathrm{s}$ (top row in each sub-figure) to target domain $T_\mathrm{t}$. Rows two to five show corresponding 3D avatar results from different viewpoints.
   }
    \label{fig:generated}
\end{figure*}

\begin{figure*}[ht]

    \centering
    \includegraphics[width=\linewidth]{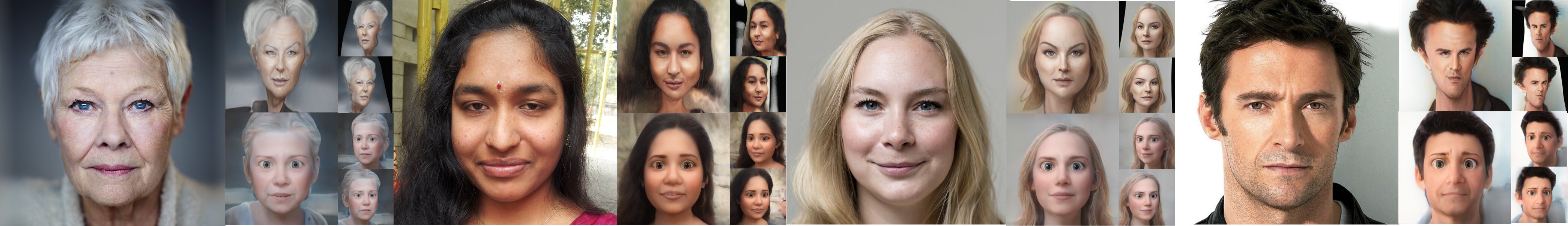}
        \caption{
    \textbf{Real to avatar results.
   } Our framework takes a single image as input and generates editable 3D Avatars. Also see Fig.~\ref{fig:real}}
    \label{fig:realtoavatar}
\end{figure*}

\section{Domain Adaptation for 3D-GANs}
\label{sec:domain}

The goal of domain adaptation for 3D-GANs is to adapt (both texture and geometry) to a particular style defined by a 2D dataset (Caricature, Anime, Pixar toons, Comic, and Cartoons~\cite{Jang2021StyleCari,yang2022Pastiche,HuoBMVC2018WebCaricature} in our case). 
In contrast to 2D-StyleGAN-based fine-tuning methods that are conceptually simpler~\cite{https://doi.org/10.48550/arxiv.2010.05334, StyleGAN2-ADA}, fine-tuning a 3D-GAN on 2D data introduces challenges in addition to domain differences, especially on maintaining the texture quality while preserving the geometry. Moreover, for these datasets, there is no explicit shape and camera information. We define the domain adaptation task as follows: Given a prior 3D-GAN \ie \red{EG3D} ($\mathrm{G_s}$) of source domain ($T_\mathrm{s}$), we aim to produce a 3D Avatar GAN ($\mathrm{G_t}$) of the target domain ($T_\mathrm{t}$) while maintaining the semantic, style, and geometric properties of $\mathrm{G_s}$, and at the same time preserving the identity of the subject between the domains ($T_\mathrm{s} \displaystyle \leftrightarrow T_\mathrm{t} $). \red{ Refer to Fig.~\ref{fig:inference} for the pipeline figure. We represent $\mathrm{G_{2D}}$ as a teacher 2D-GAN used for knowledge distillation fine-tuned on the above datasets}. Note that as $T_\mathrm{t}$ is not assumed to contain camera parameter annotations, the training scheme must suppress artifacts such as low-quality texture under different views and flat geometry (See Fig.~\ref{fig:artifacts}). In the following, we discuss the details of our method. 


\subsection{How to align the cameras?}
\label{sec:camera}

Selecting appropriate ranges for camera parameters is of paramount importance for high-fidelity geometry and texture detail. Typically, such parameters are empirically estimated, directly computed from the dataset using an off-the-shelf pose detector~\cite{EG3D}, or learned during training~\cite{imagenetGAn}. In domains we aim to bridge, such as caricatures for which a 3D model may not even exist, directly estimating the camera distribution is problematic and, hence, is not assumed by our method. Instead, we find it essential to ensure that the camera parameter distribution is consistent across the source and target domains. For the target domain, we use StyleGAN2 trained on FFHQ, fine-tuned on artistic datasets~\cite{Jang2021StyleCari,yang2022Pastiche}. Assuming that the intrinsic parameters of all the cameras are the same, we aim to match the distribution of extrinsic camera parameters of $\mathrm{G_s}$ and $\mathrm{G_{2D}}$ and train our final $\mathrm{G_t}$ using it \red{(see illustration in Fig.~\ref{fig:camera})}. To this end, we define an optimization-based method to match the sought distributions. 
%
%
The first step is to identify a canonical pose image in $G_\mathrm{2D}$, where the yaw, pitch, and roll parameters are zero. According to Karras et al., ~\cite{karras2019style}, the image corresponding to the mean latent code satisfies this property. \red{ Let $\theta$, $\phi$ be the camera Euler angles in a spherical coordinate system, $r$, $c$ be the radius of the sphere and camera lookat point, and, $\mathrm{M}$ be a function that converts these parameters into the camera-to-world matrix}. Let $I_\mathrm{s}(w, \theta, \phi, c , r) = \mathrm{G_s}(w, \mathrm{M}(\theta, \phi, c , r))$ and $I_\mathrm{2D}(w) = \mathrm{G_{2D}}(w)$ represent an arbitrary image generated by $\mathrm{G_s}$ and $\mathrm{G_{2D}}$, respectively, given the $w$ code variable. Let $k_\mathrm{d}$ be the face key-points detected by the detector $\mathrm{K_d}$~\cite{githubrepo}, then
\begin{equation}
\begin{aligned}
(c', r') :=  \argmin_{(c, r)} 
{\mathrm{L_{kd}} ( I_\mathrm{s}(w'_\mathrm{avg}, 0, 0, c , r), I_\mathrm{2D}(w_\mathrm{avg}))},    
\end{aligned}
\end{equation}
where $\mathrm{L_{kd}} (I_1, I_2) = \|{k_\mathrm{d}(I_1) - k_\mathrm{d}(I_2)}\|_1$  and $w_\mathrm{avg}$ and $w'_\mathrm{avg}$ are the mean $w$ latent codes of $\mathrm{G_{2D}}$ and $\mathrm{G_s}$, respectively. In our results, $r'$ is determined to be $2.7$ and $c'$ is approximately $[0.0,  0.05,  0.17]$. The next step is to determine a safe range of the $\theta$ and $\phi$ parameters. Following prior works, StyleFlow~\cite{abdal2020styleflow} and FreeStyleGAN~\cite{FreeStyleGAN2021} (see Fig.5 of the paper), we set these parameters as $\theta' \in [-0.45, 0.45]$ and $\phi' \in [-0.35, 0.35]$ in radians.



\subsection{What loss functions and regularizers to use?}
\label{sec:loss}
Next, although the camera systems are aligned, the given dataset may not stem from a consistent 3D model, \eg, in the case of caricatures or cartoons. This entices the generator $\mathrm{G_t}$ to converge to an easier degenerate solution with a flat geometry. Hence, to benefit from the geometric prior of $\mathrm{G_s}$, another important step is to design the loss functions and regularizers for a selected set of parameters to update in $\mathrm{G_t}$. Next, we discuss these design choices:


\noindent\textbf{Loss Functions.} To ensure texture quality and diversity, we resort to the adversarial loss used to fine-tune GANs as our main loss function. We use the standard non-saturating loss to train the generator and discriminator networks used in EG3D~\cite{chan2021efficient}. We also perform lazy density regularization to ensure consistency of the density values in the final fine-tuned model $\mathrm{G_t}$.

\noindent\textbf{Texture Regularization.} Since the texture can be entangled with the geometry information, determining which layers to update is important. To make use of the fine-style information encoded in later layers, it is essential to update the \textit{tRGB} layer parameters (outputting tri-plane features) before the neural rendering stage. \red{ \textit{tRGB} are convolutional layers that transform feature maps to 3 channels at each resolution (96 channels in triplanes)}. Moreover, since the network has to adapt to a color distribution of $T_\mathrm{t}$, it is essential to update the decoder (\textit{MLP} layers) of the neural rendering pipeline as well. Given the EG3D architecture, we also update the super-resolution layer parameters to ensure the coherency between the low-resolution and high-resolution outputs seen by the discriminator $\mathrm{D}$. 

\noindent\textbf{Geometry Regularization.} In order to allow the network to learn the structure distribution of $T_\mathrm{t}$ and at the same time ensure properties of $\mathcal{W}$ and $\mathcal{S}$ latent spaces are preserved, we update the earlier layers with regularization. This also encourages the latent spaces of $T_\mathrm{s}$ and $T_\mathrm{t}$ to be easily linked. Essentially, we update the deviation parameter $\Delta s$ from the $s$ activations of the $\mathcal{S}$ space~\cite{wu2020stylespace}. The $s$ activations are predicted by $\mathrm{A}(w)$, where $\mathrm{A}$ is the learned affine function in EG3D. \red{The $s$ activations scale the kernels of a particular layer}. In order to preserve the identity as well as geometry such that the optimization of $\Delta s$ does not deviate too far away from the original domain $T_\mathrm{s}$, we introduce a regularizer given by
\begin{equation}
\label{eq:s}
\mathrm{R}(\Delta s) :=   \|{\Delta s}\|_1.
\end{equation}
 Note that we apply $\mathrm{R}(\Delta s)$ regularization in a lazy manner, \ie, with density regularization. Interestingly, after training, we can interpolate between $s$ and $s + \Delta s$ parameters to interpolate between the geometries of samples in $T_\mathrm{s}$ and $T_\mathrm{t}$ (See Fig.~\ref{fig:s_interp}).



\noindent\textbf{Depth Regularization.} Next, we observe that even though the above design choice produces better geometry for $T_\mathrm{t}$, some samples from $\mathrm{G_t}$ can still lead to flatter geometry, and it is hard to detect these cases. We found that the problem is related to the relative depth of the background to the foreground. To circumvent this problem, we use an additional regularization where we encourage the average background depth of $\mathrm{G_t}$ to be similar to $\mathrm{G_s}$. Let $\mathrm{S_b}$ be a face background segmentation network~\cite{CelebAMask-HQ}. We first compute the average background depth of the samples given by $\mathrm{G_s}$. This average depth is given by 
\begin{equation}
a_\mathrm{d} := \frac{1}{M} \sum_{n=1} ^{M} (\frac{1}{N_n} \|D_n \odot \mathrm{S_b}(I_n)\|_F^2).
\end{equation}
Here, $D_n$ is the depth map of the image $I_n$ sampled from $G_\mathrm{s}$, $\odot$ represents the \textit{Hadamard} product, $M$ is the number of the sampled images, and $N_n$ is the number of background pixels in $I_n$. Finally, regularization is defined as:
\begin{equation}
\label{eq:depth}
\mathrm{R}(D) :=  \|a_\mathrm{d} \cdot J - (D_\mathrm{t} \odot \mathrm{S_b}(I_\mathrm{t}))\|_F,   
\end{equation}
where $D_\mathrm{t}$ is the depth map of the image $I_\mathrm{t}$ sampled from $\mathrm{G_t}$ and $J$ is the matrix of ones having the same spatial dimensions as $D_\mathrm{t}$.

\begin{figure}[t]
    \centering
    \includegraphics[ width=\linewidth]{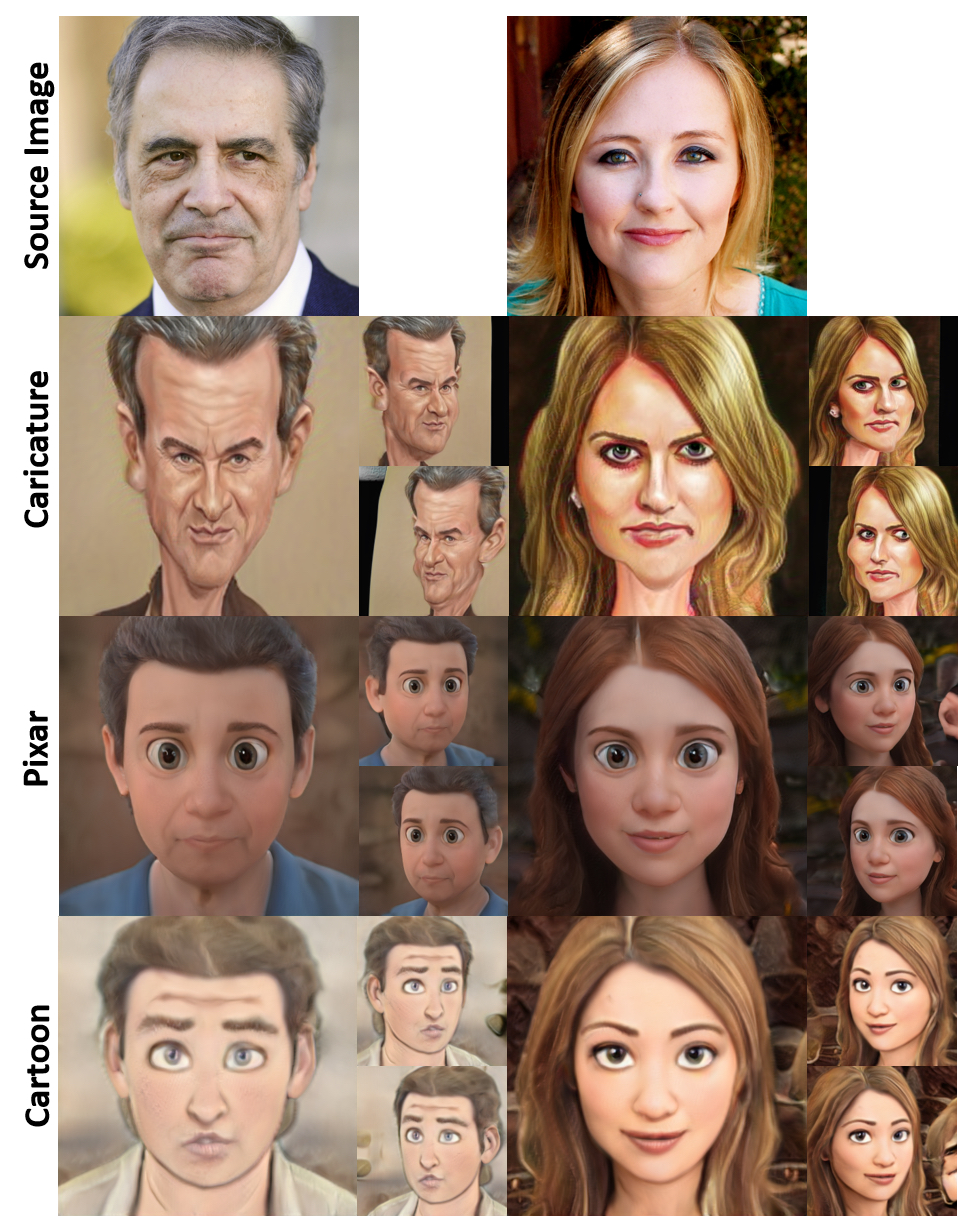}
    \vspace{-3mm}
    \caption{
    \textbf{3D avatars from real images.} Projection of real images on the 3D avatar generators. 
   }
   \vspace{-5mm}
    \label{fig:real}
\end{figure}




\subsection{What discriminator to use?}

Given that the data in $T_\mathrm{s}$ and $T_\mathrm{t}$ is not paired and $T_\mathrm{t}$ is not assumed to contain camera parameter annotations, the choice of the discriminator ($\mathrm{D}$) used for this task is also a critical design choice. Essentially, we use the unconditional version of the dual discriminator proposed in EG3D, and hence, we do not condition the discriminator on the camera information. As a result, during the training, $\mathrm{G_t}$ generates arbitrary images with pose using $\mathrm{M}(\theta', \phi', c', r')$, and the discriminator discriminates these images using arbitrary images from $T_\mathrm{t}$. We train the discriminator from scratch and in order to adapt $T_\mathrm{s} \rightarrow T_\mathrm{t}$, we use the StyleGAN-ADA~\cite{Karras2020ada} training scheme and use $\mathrm{R1}$ regularization.

\subsection{How to incorporate larger geometric deformations between domains?}
\label{sec: tps}

While the regularizers are used to limit the geometric changes when adapting from $T_\mathrm{s}$ to $T_\mathrm{t}$, modeling large geometric deformations, e.g., in the caricature dataset is another challenge. One choice to edit the geometry is to use the properties of tri-plane features learned by EG3D. We start out by analyzing these three planes in $\mathrm{G_s}$. We observe that the frontal plane encodes most of the information required to render the final image. To quantify this, we sample images and depth maps from $\mathrm{G_s}$ and swap the front and the other planes from two random images. Then we compare the difference in \textit{RGB} values of the images and the \textit{Chamfer} distance of the depth maps. While swapping the frontal tri-planes, the final images are completely swapped, and the \textit{Chamfer} distance changes by $80 \sim 90 \%$ matching the swapped image depth map. In the case of the other two planes, the \textit{RGB} image is not much affected and the \textit{Chamfer} distance of the depth maps is reduced by only $20 \sim 30 \%$ in most cases. 

Given the analysis, we focus to manipulate the \textit{2D} front plane features to learn additional deformation or exaggerations.  We learn a \textit{TPS (Thin Plate Spline)}~\cite{githubrepotps} network on top of the front plane. Our \textit{TPS} network is conditioned both on the front plane features as well as the $\mathcal{W}$ space to enable multiple transformations. The architecture of the module is similar to the standard StyleGAN2 layer with an \textit{MLP} appended at the end to predict the control points that transform the features.  Hence, as a byproduct, we also enable 3D-geometry editing guided by the learned latent space. We train this module separately after $\mathrm{G_t}$ has been trained. We find that joint training is unstable due to exploding gradients arising from the large domain gap between $T_\mathrm{s}$ and $T_\mathrm{t}$ in the initial stages. Formally, we define this transformation as:
\begin{equation}
\begin{aligned}
\mathrm{T}(w, f) : = \Delta c,
\end{aligned}
\end{equation}
where, $w$ is the latent code, $f$ is the front plane, and $c$ are the control points. 



 Let $c_\mathrm{I}$ be the initial control points producing an identity transformation, $(c_1,c_2)$ be the control points corresponding to front planes $(f_1,f_2)$ sampled using $\mathcal{W}$ codes $(w_1,w_2)$, respectively, and $(c'_1,c'_2)$ be points with $(w_1,w_2)$ swapped in the \textit{TPS} module. To regularize and encourage the module to learn different deformations, we have
\begin{equation}
\begin{aligned}
\mathrm{R}(\mathrm{T}_1) := \alpha  \sum_{n=1} ^{2} \| c_I - c_n \|_1 
- \beta \|  c_1 - c_2 \|_1  - \sigma  \| c'_1 - c'_2 \|_1.
\end{aligned}
\label{eq:6}
\end{equation}

\red{We use initial control point regularization to regularize large deviations in the control points which would otherwise explode.}
Additionally, to learn extreme exaggerations in $T_\mathrm{t}$ and \red{\textit{‘in expectation’}, conform to the target distribution in the dataset}, we add an additional loss term. Let $\mathrm{S}(I)$ be the soft-argmax output of the face segmentation network~\cite{CelebAMask-HQ} given an image $I$ and assuming that $\mathrm{S}$ generalizes to caricatures, then
\begin{equation}
\begin{aligned}
\mathrm{R}(\mathrm{T}_2) :=  \| \mathrm{S}(\mathrm{G_t}(w)), \mathrm{S}(I_\mathrm{t}) \|_1
\end{aligned}
\label{eq:7}
\end{equation}

\red{Eq.~\ref{eq:6}, Eq.~\ref{eq:7}, and adversarial training loss are used to train the \textit{TPS} module. We adopt gradient clipping to make sure that the training does not diverge. See the illustrations in Fig.~\ref{fig:tps_inference} and Fig.~\ref{fig:inference}.}

\begin{figure}[t]
    \centering
    \includegraphics[width=\linewidth]{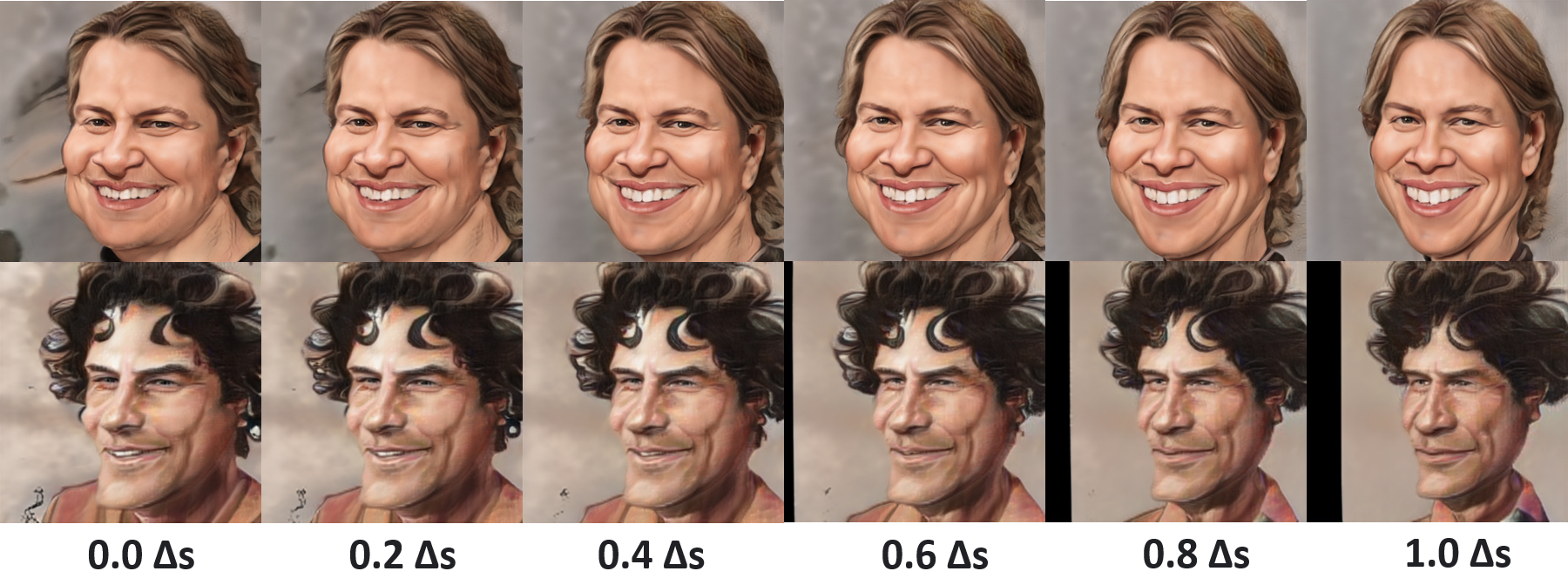}
    \vspace{-4mm}
    \caption{
    \textbf{Interpolation of $\Delta s$.} Geometric deformation using the interpolation of learned $\Delta s$ parameters.
   }
   \vspace{-4mm}
    \label{fig:s_interp}
\end{figure}

\begin{figure*}[t]
    \centering
    \includegraphics[width=\linewidth]{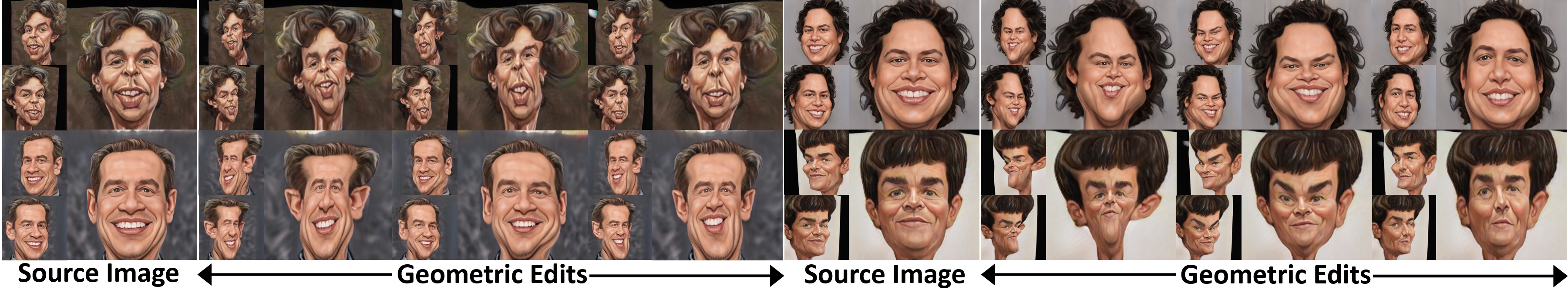}
    \vspace{-5mm}
    \caption{
    \textbf{Deformations using \textit{TPS}.} Geometric edits using our proposed \textit{TPS (Thin Plate Spline)} module learned on the frontal tri-plane features. Each sub-figure shows a 3D avatar and three examples of \textit{TPS} deformations sampled from the learned 3D deformation space. 
   }
   \vspace{-3mm}
    \label{fig:tps}
\end{figure*}

\section{Personalized Avatar Generation and Editing}
\label{sec:personalized_avatar}


Although 3D domain adaptation adapts $T_\mathrm{s} \displaystyle \leftrightarrow T_\mathrm{t}$, it is still a challenge to effectively link the latent spaces of $\mathrm{G_s}$ and $\mathrm{G_t}$ to generate personalized 3D avatars using a single photograph as the reference image. Particularly, the challenge arises due to the discrepancy in the coupled latent spaces when dealing with the projection of real photographs on 3D generators. Moreover, one would like to edit and animate these 3D avatars. \\
\noindent\textbf{Projection.} The task is to project a real image into the latent space of $\mathrm{G_s}$, transfer the latent to $\mathrm{G_t}$, and further optimize it to construct a 3D avatar. First, we use an optimization-based method to find the $w$ code that minimizes the similarity between the generated and the real image in $\mathrm{G_s}$. To achieve this, the first step is to align the cameras. We follow the steps mentioned in Sec.~\ref{sec:camera} for this step. Next, we use pixel-wise \textit{MSE} loss and \textit{LPIPS} loss to project the image into $\mathrm{G_s}$~\cite{abdal2019image2stylegan}. Additionally, to preserve the identity of the subject, we use attribute classifiers \eg caricature dataset~\cite{HuoBMVC2018WebCaricature} provides the coupled attribute information of real images and caricatures. We use such attribute classifier~\cite{HuoBMVC2018WebCaricature, Jang2021StyleCari} in a post-hoc manner as we notice that such networks can affect the texture in the target domain and could degenerate to narrow style outputs if applied during training. Moreover, such networks may not be available for all target domains. To avoid overfitting into $\mathrm{G_s}$ and encourage the easier transfer of the optimized latent code to $\mathrm{G_t}$, we use $\mathcal{W}$ space optimization for this step. Finally, we initialize this $w$ code for $\mathrm{G_t}$ and use additional attribute classifier loss~\cite{Jang2021StyleCari} for $T_\mathrm{t}$ domain along with depth regularization $\mathrm{R}(D)$ (Eq.~\ref{eq:depth}). As an approximation, we assume that attribute classifier~\cite{HuoBMVC2018WebCaricature, Jang2021StyleCari} generalizes across all domains.
 We use $\mathcal{W}$/$\mathcal{W+}$ space optimization to control the quality and diversity of the outputs.
 
 Let $x$ be the source image, let pixel-wise $MSE$ loss be represented as $L_{mse}(x,w, G) = MSE(x, \mathrm{G}(w, \mathrm{M}(\theta', \phi', c', r')))$, and let $LPIPS$ loss be represented as $L_{lpips}(x,w, G) = LPIPS(x, \mathrm{G}(w, \mathrm{M}(\theta', \phi', c', r')))$ where camera parameters are determined by Sec.~\ref{sec:camera}. Let $L_{d}(w)$ be the depth regularizer (Eq.~\ref{eq:depth}) and $A_t(x,w, G)$ be the attribute classifier loss. We define the algorithm of projection of a single source image into 3D avatars in Algorithm~\ref{alg:real}.

 \begin{algorithm}[ht]
\SetAlgoLined
 \KwIn{source image $x \in \mathbb{R}^{n \times n \times 3}$; $\mathrm{G_s}$, $\mathrm{G_t}$, gradient-based optimizer $F'$ and $F''$.}
 \KwOut{the embedded code $w'$ for $\mathrm{G_t}$}
 Initialize() the code $w$ = $w_{avg}$\;
 \While{not converged}{
  $L \leftarrow L_{mse}(x,w, \mathrm{G_s}) + L_{lpips}(x,w, \mathrm{G_s}) + L_{d}(w) + A_t(x,w, \mathrm{G_s}) $\;
  $w \leftarrow w - \eta F'(\nabla_w L,w)$\;
  }
 Initialize() the code $w'$ = $w$\;
 
 \While{not converged}{
  $L' \leftarrow L_{mse}(x,w', \mathrm{G_t}) + L_{lpips}(x,w', \mathrm{G_t}) + L_{d}(w') + A_t(x,w', \mathrm{G_t})$\;
  $w' \leftarrow w' - \zeta F''(\nabla_{w'} L',w')$\;
 }
 \caption{Projection of single image into 3D Avatar.}
 \label{alg:real}
\end{algorithm}

\noindent\textbf{Editing and Animation.} Since our 3D domain adaptation is designed to preserve the properties of $\mathcal{W}$ and $\mathcal{S}$ spaces, we can perform semantic edits via InterFaceGAN~\cite{shen2020interfacegan}, GANSpace~\cite{harkonen2020ganspace}, StyleSpace~\cite{wu2020stylespace} \etc, and geometric edits using \textit{TPS} (Sec.~\ref{sec: tps}) and $\Delta s$ interpolation (Sec.~\ref{sec:loss}). To perform video editing, we design an encoder for EG3D based on \textit{e4e}~\cite{tov2021designing} to encode videos and transfer the edits from $\mathrm{G_s}$ to $\mathrm{G_t}$ based on the $w$ codes~\cite{https://doi.org/10.48550/arxiv.2205.13996, DBLP:journals/corr/abs-2201-13433, DBLP:journals/corr/abs-2201-08361}. We leave a more fine-grained approach for video processing as future work. 




%% file: 4-experiment.tex
\section{Results}

\subsection{Quantitative Results}
\label{sec:quant}
In this section, we consider three important evaluations to verify the quality of the texture, geometry, and identity preservation in the new domain using the Caricature, Cartoons, and Pixar toons datasets. We also evaluate the ablation of our design choices and conduct a user study to assess the quality of generated avatars. In the evaluation, let $\mathrm{G_{base}}$ be the baseline na\"ive fine-tuning method which is trained with all the parameters using the losses in EG3D fine-tuned from FFHQ trained prior $\mathrm{G_s}$. Note here we still align the cameras in $\mathrm{G_{base}}$ using the method defined in Sec.~\ref{sec:camera} and use adaptive discriminator~\cite{Karras2020ada} with $\mathrm{R1}$ regularization for a fair comparison.

\noindent\textbf{Texture Quality.} To verify the quality of the texture, diversity of samples as well as to some extent, the geometry in the target domain $T_\mathrm{t}$, we compare the \textit{FID}~\cite{FiD} scores using $\mathrm{G_{base}}$ and $\mathrm{G_t}$ in Table~\ref{tab:fid}. Note that in the case of Caricatures, we report two scores \ie with and without using the attribute classifier loss in the training as discussed in Sec.~\ref{sec:personalized_avatar}. Notice that our method outperforms the na\"ive  baseline method by a huge margin in some cases, especially in Caricatures and Cartoons. We attribute these differences to the mode collapse prone training of $\mathrm{G_{base}}$ which is correlated with flat geometry degenerate solution. We show visual results of the flat geometries learned by $\mathrm{G_{base}}$ and comparison in Fig.~\ref{fig:artifacts}.

\noindent\textbf{Geometric Quality.} To quantify the flat geometries, in Table~\ref{tab:geometry}, we show three scores that help us understand such degenerate solutions. Here we consider coupled depth maps generated from sampling in the domains $T_\mathrm{s}$ ($\mathrm{G_s}$) and $T_\mathrm{t}$ ($\mathrm{G_t}$ and $\mathrm{G_{base}}$). First, we compute the expectation of the absolute mean differences ($M_\mathrm{d}$) of the corresponding foreground depth maps sampled from $T_\mathrm{s}$ and $T_\mathrm{t}$. We also compute the expectation of the absolute standard deviation differences ($S_\mathrm{d}$) for the same setting. Here, we assume that the flatter geometries have a large difference in the depth maps as compared to the prior as indicated by $M_\mathrm{d}$. Moreover, $S_\mathrm{d}$ computes the distance in the distribution of the depth values, where a larger difference indicates a narrow distribution, and hence a flatter geometry. We also notice that the flat geometry is correlated with the generator learning diverse poses when images are rendered under standard canonical camera parameters \ie $\mathrm{M}(0, 0, c, r)$. We hypothesize in the case of the flatter geometries, the model learns to pose information in the earlier layers instead of being camera view-dependent. To quantify this, since pose information may not be available for some domains (\eg cartoons), we compute the $\mathrm{R}(\mathrm{T_2})$ scores between corresponding images in the domain $T_\mathrm{s}$ ($\mathrm{G_s}$) and $T_\mathrm{t}$ ($\mathrm{G_t}$ and $\mathrm{G_{base}}$). Note that these scores are computed without the \textit{TPS} module. Our scores are lower in all three metrics, hence, validating that our method avoids the degenerate solution and preserves the geometric distribution of the prior.

\input{tables/tab_fid}

\input{tables/tab_geo}

\input{tables/tab_identity}
\noindent\textbf{Identity Preservation.} Identity preservation score is another important evaluation to check the quality of latent space linking between $\mathrm{G_s}$ and $\mathrm{G_t}$. In Table~\ref{tab:identity}, we compute the attribute loss (\textit{BCE} loss) between the domains $T_\mathrm{s}$ and $T_\mathrm{t}$ using the attribute  classifiers~\cite{Jang2021StyleCari,HuoBMVC2018WebCaricature}. Note that our method is able to preserve the identity better across the domains.

\begin{figure}[t]
    \centering
    \includegraphics[width=\linewidth]{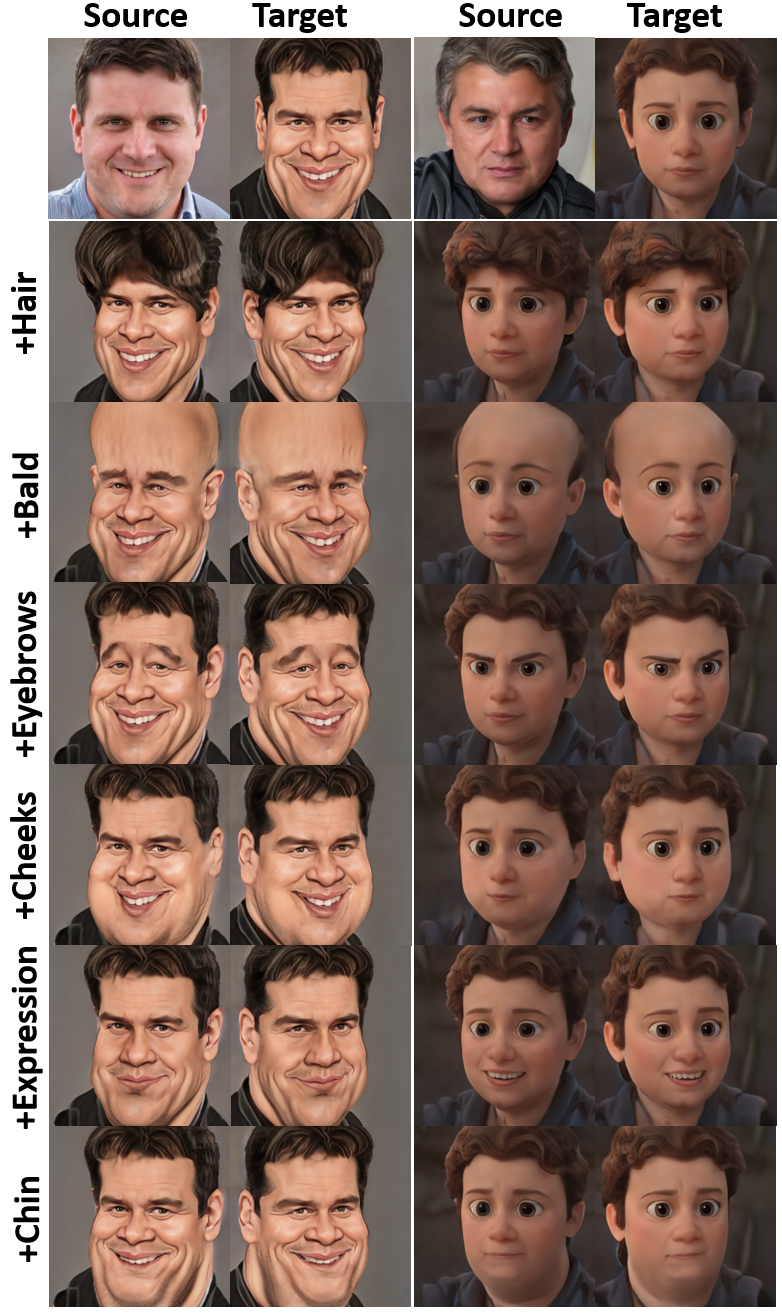}
    \caption{
    \textbf{Local edits.} Local edits performed on the 3D avatars using the $S$ space.
   }
   \vspace{-5mm}
    \label{fig:edits}
\end{figure}

\begin{figure*}[ht!]
    \centering
    \includegraphics[width=\linewidth]{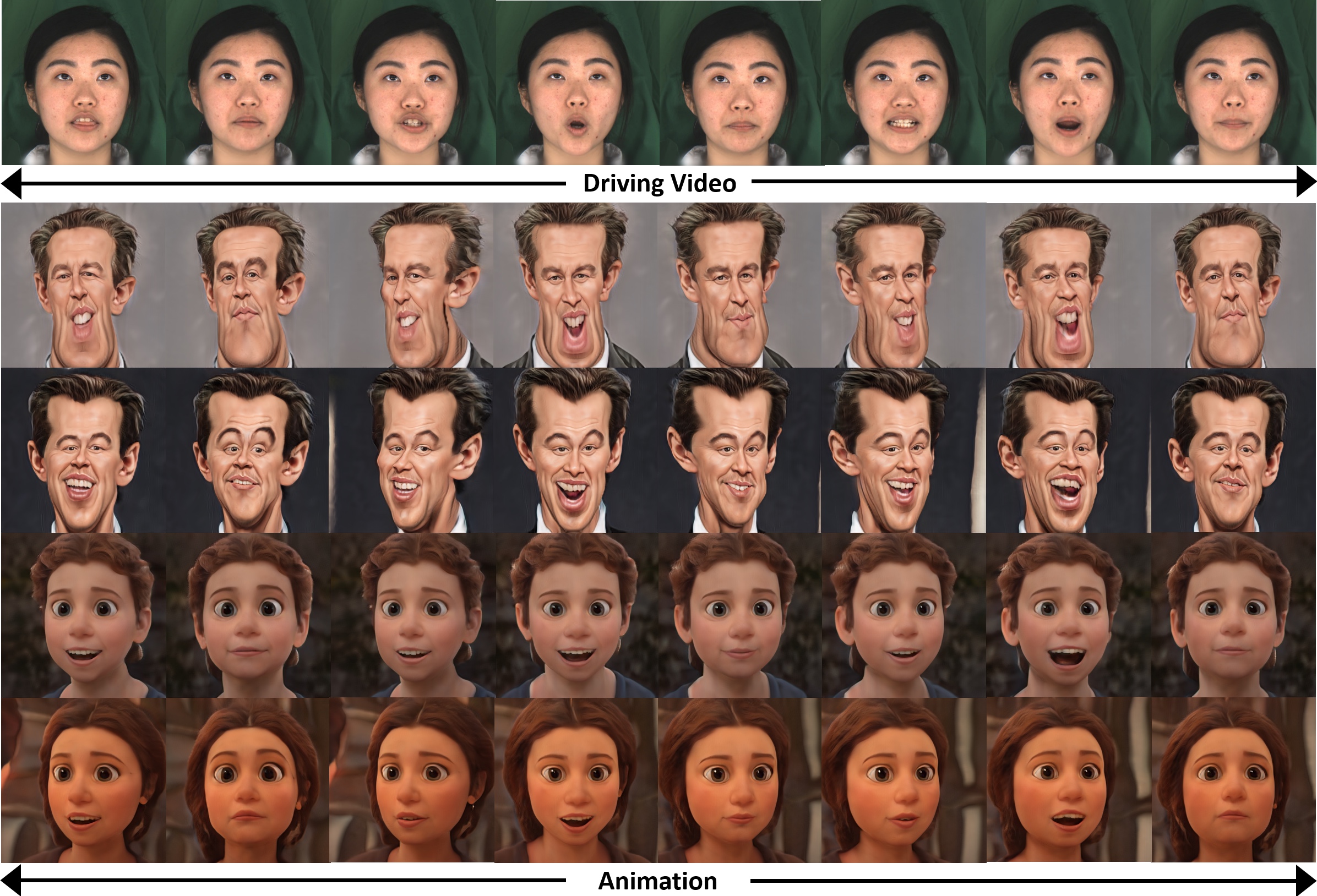}
    \caption{
    \textbf{3D avatar animation.} Animation of 3D avatars generated using a driving video encoded in source domain $T_s$ and applied to samples in target domain $T_t$. The top row shows the driving video and the subsequent rows show generated animations using a random Caricature or Pixar toon. The head pose is changed in each frame of the generated animation to show 3D consistency. 
   }
   \vspace{-3mm}
    \label{fig:video}
\end{figure*}

\noindent\textbf{Ablation Study.} In order to validate the importance of the losses, and components of our design choices, in Table~\ref{tab:ablation}, we show an ablation study of these components with regularizers. Note that we evaluate these design choices on the caricature dataset. Notice adding each component improves the corresponding scores in FID, $M_d$, $S_d$, and $ID$ as discussed. Notice that by adding the $TPS$ module, the FID is still comparable to $\mathrm{G_{t}}$. The slight drop is attributed to the stretching and squeezing of some parts of the texture (See Fig.~\ref{fig:tps}). Nevertheless, by adding this module, we achieve better control over geometry and produce exaggerated features for a small drop in texture quality. In order to show that the $TPS$ transformations are not random, we compute the FID scores with randomly perturbed front plane features which are derived from the perturbations of control points near the face \eg taken at the early stages of training after it has stabilized a bit. This setup has less perturbation in the background. We found that the FID score is worse \ie  $25.5$ hence validating non-random transformations. 

\input{tables/ablation.tex}

The purpose of adding the \textit{TPS} module is to model the exaggerated geometries in the data and at the same time achieve geometric editing and animation capabilities (see \href{https:/rameenabdal.github.io/3DAvatarGAN}{Project Page}). We ablate (see Table~\ref{tab:kp_variation}) the average face Keypoint Distance between the paired FFHQ-generated images and corresponding avatars and the face Keypoint Variation (standard deviation of keypoints) with and without the \textit{TPS} module. The results show that with \textit{TPS} module, the deviations are large and match the exaggerations. 
\input{tables/scores}

\noindent\textbf{User Study.} For results on real images refer to Fig.~\ref{fig:realtoavatar}. We conducted a user study using 50 real images (25 caricatures, and 25 Pixar) on identity preservation and 3D consistency versus the baseline method. We asked 21 unique workers where each triplet was reviewed by 10 workers and our avatars were chosen \textbf{92\%} of the time.

\subsection{Qualitative Results}
\label{sec:qual}
For qualitative results, we show the results of the domain adaptation, as well as the personalized edits (geometric and semantic), performed on the resultant 3D avatars. First, in order to show the quality of domain adaptation, identity preservation, and geometric consistency, in Fig.~\ref{fig:generated}, we show results from $\mathrm{G_s}$ and corresponding results from 3D avatar generator $\mathrm{G_t}$ trained on Caricatures, Pixar toons, Cartoons, and Comic domains. Next, in order to show that the method generalizes to real images, we use the method described in Sec.~\ref{sec:personalized_avatar} to project and transfer the latent code from $\mathrm{G_s}$ to $\mathrm{G_t}$ to produce the 3D avatars. In Fig.~\ref{fig:real}, we show our results of real to 3D avatar transfer. Notice the quality both in terms of texture as well as geometry for both these results achieved by our method. Next, we show geometric and semantic edits possible to produce personalized 3D avatars:\\

\vspace*{-2mm}
\noindent\textbf{Geometry Edits.} We show two type of geometric edits \ie $\Delta s$ interpolation (Sec.~\ref{sec:loss}) and deformation using \textit{TPS} (Sec.~\ref{sec: tps}). First, in Fig.~\ref{fig:s_interp}, we show the geometry interpolation by interpolating between original $s$ activations of $\mathrm{G_s}$ and learned $\Delta s$ parameters. In Fig.~\ref{fig:tps}, we show some additional exaggerations in caricatures using the learned 3D deformation latent space of \textit{TPS} module.

\noindent\textbf{Semantic Edits and Animation.} Since in our method, we encourage the latent regularization to preserve the properties of the latent space learned by the $\mathrm{G_s}$ generator, in Fig.~\ref{fig:edits} we show $\mathcal{S}$ space edits performed on the 3D avatars. Notice the quality of edits in terms of locality and adaptability. Additionally, we can edit semantics like hair as opposed to 3D morphable model based methods. In Fig.~\ref{fig:video}, thanks to the latent space semantics preservation ensured by our method, we can perform some video edits to create a coherent animation based on the difference of $w$ codes of video encoded in $\mathrm{G_s}$ (Sec.~\ref{sec:personalized_avatar}) and applied to layers $7 - 10$ in $\mathrm{G_t}$. Notice the quality of expressions, identity preservation, and 3D consistency across each identity in each row.

%% file: tables/tab_fid.tex
\begin{table}%
\caption{
\textbf{FID Computation.}
FID (Fréchet Inception Distance) between the 2D dataset and the samples generated by the fine-tuned 3D GAN using baseline ($\mathrm{G_{base}}$) and Ours ($\mathrm{G_t}$). '*' represents the score with the inclusion of the attribute classifier loss discussed in Sec.~\ref{sec:loss}.}
\label{tab:fid}
\begin{minipage}{\columnwidth}
\begin{center}
\begin{tabular}{rlll}

  \toprule
  Method  & Caricatures   &  Cartoons   &  Pixar Toons\\ \midrule
$\mathrm{G_{base}}$   & 67.8  &79.0 & 15.1 \\
$\mathrm{G_t}$ (Ours)    & \textbf{19.4/20.2*}  & \textbf{12.8} & \textbf{12.4} \\

  \bottomrule
\end{tabular} 
\end{center}
\end{minipage}
\end{table}%

%% file: tables/tab_geo.tex
\begin{table}%

\def\ua{\ensuremath{\uparrow}}
\def\da{\ensuremath{\downarrow}}
\caption{
\textbf{Geometry Evaluation.}
Comparing the geometry using baseline method ($\mathrm{G_{base}}$) and Ours ($\mathrm{G_t}$). For the definition of $M_\mathrm{d}$, $S_\mathrm{d}$ and $\mathrm{R}(\mathrm{T_2})$, refer to Sec.~\ref{sec:quant}. }
\label{tab:geometry}
\begin{minipage}{\columnwidth}
\begin{center}
\begin{tabular}{rllll}

  \toprule
  Metric & Method  & Caricatures   &  Cartoons   &  Pixar \\ \midrule
  $M_\mathrm{d}$ \da & $\mathrm{G_{base}}$ &0.47  & 0.21  & 0.29 \\
 &  $\mathrm{G_t}$ (Ours) &\textbf{0.21}   &\textbf{0.13}   &\textbf{0.13} \\
 \midrule
$S_\mathrm{d}$ \da &$\mathrm{G_{base}}$ &0.22 &0.14 &0.15 \\
& $\mathrm{G_t}$ (Ours) &\textbf{0.15 } &\textbf{0.10 } &\textbf{0.09}
\\
\midrule

$\mathrm{R}(\mathrm{T_2})$ \da &$\mathrm{G_{base}}$ &2.99 &3.39 &4.01 \\
& $\mathrm{G_t}$ (Ours) &\textbf{2.27 } &\textbf{1.62 } &\textbf{1.56}
\\
  \bottomrule
\end{tabular} 
\end{center}
\end{minipage}
\end{table}%

%% file: tables/tab_identity.tex
\begin{table}%
\caption{
\textbf{Identity Preservation.}
Identity preservation using baseline ($\mathrm{G_{base}}$) and Ours ($\mathrm{G_t}$).}
\label{tab:identity}
\begin{minipage}{\columnwidth}
\begin{center}
\begin{tabular}{rlll}

  \toprule
  Method  & Caricatures   &  Cartoons   &  Pixar Toons\\ \midrule
$\mathrm{G_{base}}$   & 1.28  &0.92 & 0.85 \\
$\mathrm{G_t}$ (Ours)    & \textbf{0.87}  & \textbf{0.81} & \textbf{0.73} \\

  \bottomrule
\end{tabular} 
\end{center}
\end{minipage}
\end{table}%

%% file: tables/ablation.tex
\begin{table}%
\caption{
\textbf{Ablation.}
Ablation of the design choices made in Sec. \ref{sec:domain}. $\mathrm{cam}$ stands for the analysis in Sec. \ref{sec:camera}, $\mathrm{Reg}$ stands for the model after applying Eq. \ref{eq:s}, $\mathrm{DReg}$ stands for the model after applying Eq. \ref{eq:depth}, and $\mathrm{TPS}$ stands for the model after applying Eq. 5 - 7. This is the $\mathrm{G_{t}}$ used in the comparison. Note that by adding $\mathrm{TPS}$ the scores are affected as the geometry is exaggerated \eg the identity is affected. This module can be added to do geometry editing.}
\label{tab:ablation}
\begin{minipage}{\columnwidth}
\begin{center}
\begin{tabular}{rlllll}

  \toprule
  Method  & FID &  $M_d$   & $S_d$  & $ID$\\ \midrule
$\mathrm{G_{base}}$ - $cam$  &90.8 &0.47 &0.33 &1.348\\
$\mathrm{G_{base}}$ &67.8 &0.47 &0.22 &1.272 \\
+ $\mathrm{Reg}$ &19.0 &0.22 &0.22 &0.889 \\
+ $\mathrm{DReg}$ &19.4 &0.21 &0.15 &0.879 \\
+ $\mathrm{TPS}$ &20.6 &0.25 &0.20 &0.924 \\
  \bottomrule
\end{tabular} 
\end{center}
\end{minipage}
\end{table}%

%% file: tables/scores.tex
\begin{table}%
\vspace{-1mm}
\caption{ 
\textbf{Ablation of TPS on metrics based on facial keypoints.}}
\vspace*{-3mm}
\label{tab:kp_variation}
\begin{minipage}{\columnwidth}
\begin{center}
\begin{tabular}{rll}

  \toprule
  Metric  & with TPS     &  without TPS\\ \midrule
Avg. Keypoint Distance   & \textbf{5.7}  &4.3 \\
Avg. Keypoint Variation   & \textbf{0.06} & 0.04 \\

  \bottomrule
\end{tabular} 
\end{center}
\end{minipage}
\end{table}%

%% file: 5-conclusion.tex
\section{Conclusion}

We tackled two open research problems in this paper. In the first part, we proposed the first domain adaptation method for 3D-GANs to the best of our knowledge. This part yields two linked EG3D generators, one in the photo-realistic source domain of faces, and another EG3D generator in an artistic target domain. As possible target domains, we show results for cartoons, caricatures, and comics. In the second part, we built on domain adaptation to create 3D avatars in an artistic domain that can be edited and animated. Our framework consists of multiple technical components introduced in this paper. First, we propose a technique for camera space estimation for artistic domains. Second, we introduce a set of regularizers and loss functions that can regularize the fine-tuning of EG3D in such a way that enough of the 3D structure and geometry of the original model is kept, while the distinguishing attributes of the artistic domain, such as textures and colors and local geometric deformations can still be learned. Third, we introduce a geometric deformation module that can reintroduce larger geometric deformations in a controlled manner. These larger geometric deformations can interact and cooperate with EG3D so that semantic edits are still possible. Finally, we propose an embedding algorithm that is especially suitable for two linked EG3D generator networks.


%% file: suppl.tex
 \begin{figure*}[t!]
    \centering
    \includegraphics[width=\linewidth]{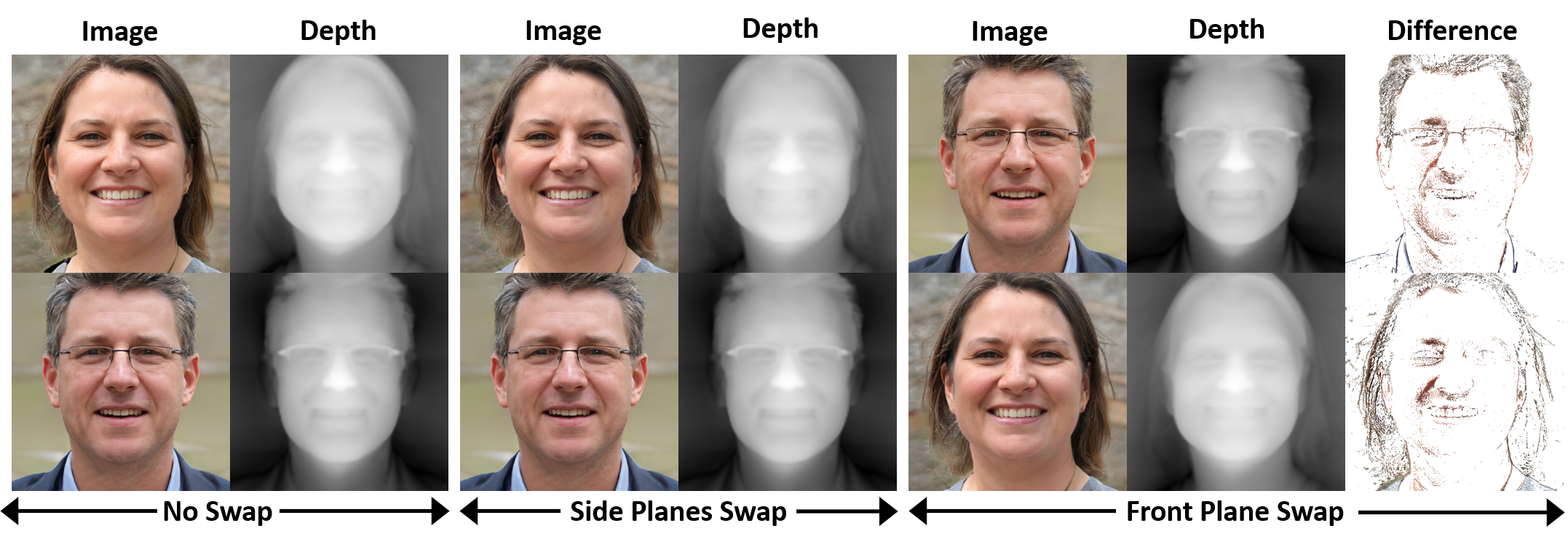}
    \caption{
    \textbf{Swaping tri-planes.} Validation of the information stored in the front tri-plane. Given two images and their tri-plane representations, there is almost no change in the final output if the side planes are swapped. While the output changes completely when the front tri-plane is swapped. 
   }
    \label{fig:swap}
\end{figure*}

\section{Limitations}
Our method also has some limitations. Overall, the visual quality is limited by the quality of StyleGAN2 pretraining. While we found the quality to be very high for the datasets shown in the paper, it relies on hundreds of images in the target domain to be available. Would be interesting to do few- shot domain adaptation in the future.
Further, edits are largely limited to semantic edits of EG3D and global space deformations by \textit{TPS}. Our method does not enable fine-grained geometric edits.
Finally, a large part of our method is face specific. We justify this specialization by the importance of human models and the specific target domain of editable 3D avatars. We nevertheless believe that domain adaptation of general 3D-GANs will be an interesting avenue of future work. 
\section{Ethical Concerns}
Deep learning-based image and video processing is a tool for image/video understanding, animation, and artistic expression. Similar to most software in this domain, our work could be used to produce offensive results. An application of concern would be if a user would generate offensive caricatures and cartoons of other people without consent. This can be used to insinuate biases against people or can have a detrimental effect on a person's autonomy, dignity, and/or privacy. The images used in this work are taken or derived from the FFHQ~\cite{githubrepoffhq} dataset which has appropriate licenses for non-commercial research use and to the best of our knowledge, is committed to protecting the privacy of the individuals who do not wish to be included. 

\section{Training Details}

We train our models on 4 V100 GPUs with a batch size of 8. Similar to EG3D, we start training from the neural rendering resolution of  $64^2$ which is increased during the training. Then we do fine-tuning on $128^2$ resolution to produce the final $512^2$ outputs. We sample $100k$ samples from each dataset. We train Caricatures on $\sim$ $880$ $kimgs$, Pixar, and Cartoons on $\sim$ $500$ $kimgs$. We fine-tune these models on $128^2$ neural rendering resolution for an additional $80 - 160$ $kimgs$. Other hyperparameters of learning rates are the same as the EG3D. We train the \textit{TPS} module on $\sim$$2000$ $kimgs$. We set the weight for the regularization term $\mathrm{R}(\Delta s)$ as $0.001$, and $\mathrm{R}(D)$ as $0.005$. For the \textit{TPS} training we use the weights for $\alpha$, $\beta$, $\sigma$ and $\mathrm{R}(\mathrm{T}_2)$ as $150$, $1$, $3$ and $1$ respectively. For inversion, we perform $200$ steps for the source domain inversion and $400$ steps for the target domain to generate the final avatar.


\red{\section{Comparison to 2D-GANs} The quality drop is expected when the final model is compared with 2D-GANs as we derive our datasets from these 2D-GANs fine-tuned on avatar datasets. In Table~\ref{tab:2d_comp}, we compute the FID scores between the datasets (size $\sim$200 images for DualStyleGAN dataset) used to train these 2D-GANs and our corresponding avatar generators (size $\sim$10k images). We used the \textit{3DCaricShop}~\cite{qiu20213dcaricshop} dataset for Caricatures as the authors of \textit{WebCaricature} did not reply with the download link. The scores are comparable, even better in the case of Caricatures and Pixar, probably due to our regularizers including 3D view consistency.}

\section{Importance of Front Tri-plane Features}

 As discussed in Sec. 3.4 of the main paper, the front tri-plane of the EG3D architecture encodes most of the texture and depth information in the output. In Fig.~\ref{fig:swap}, we show two images with their front and other side plane swapped. Then we show the corresponding effect on the output image. Notice that the results are consistent with the analysis in Sec. 3.4 where the front tri-plane dominates the information for output texture and depth.

\section{Stylization} 

\begin{figure}[t]
    \centering
    \includegraphics[width=\linewidth]{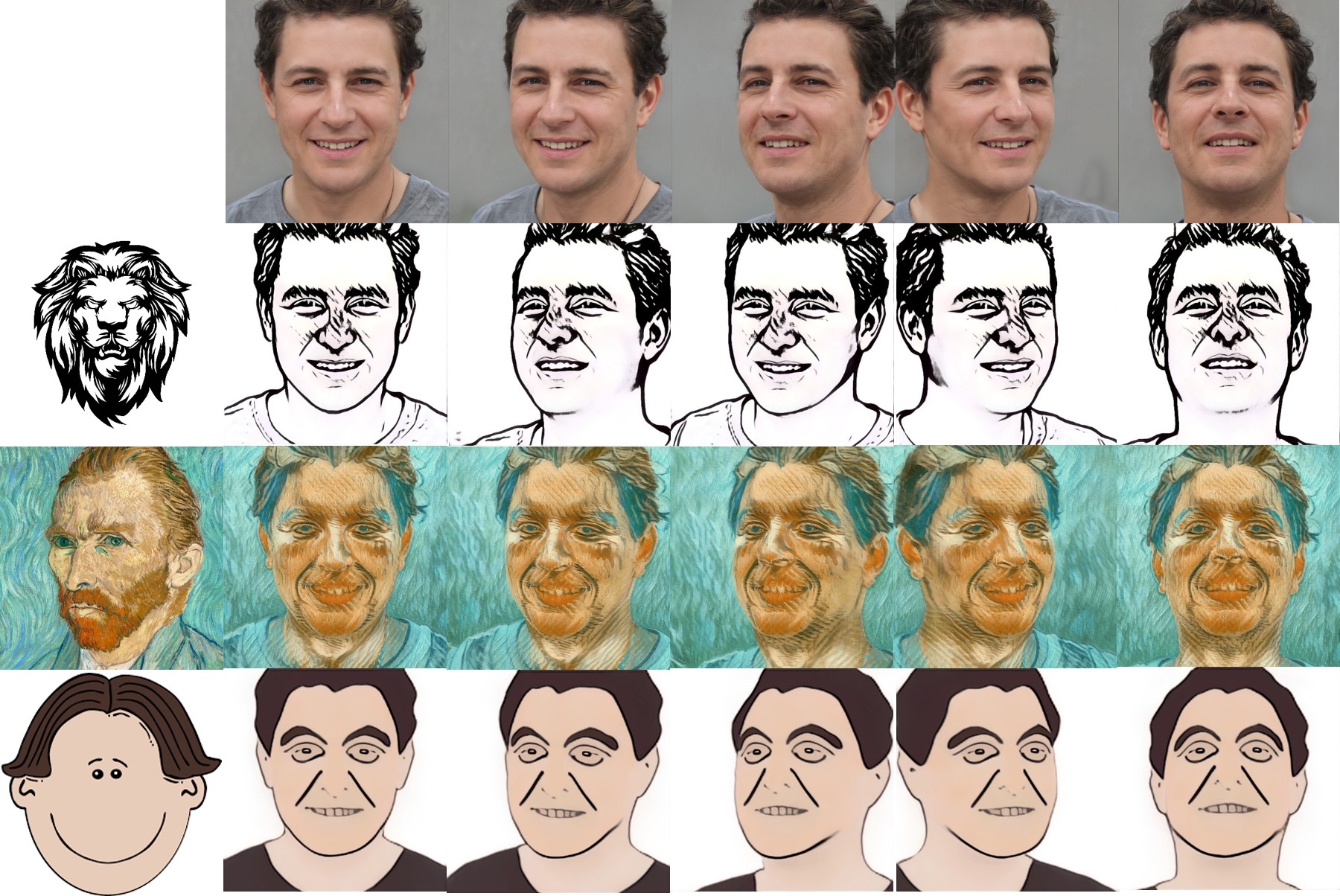}
    \caption{
    \textbf{Validation of texture regularization.} Style Transfer using the texture layers discussed in the main paper. Here the style of the image is changed using the texture layers.
   }
    \label{fig:texture}
\end{figure}

\begin{figure}[t]
    \centering
    \includegraphics[width=\linewidth]{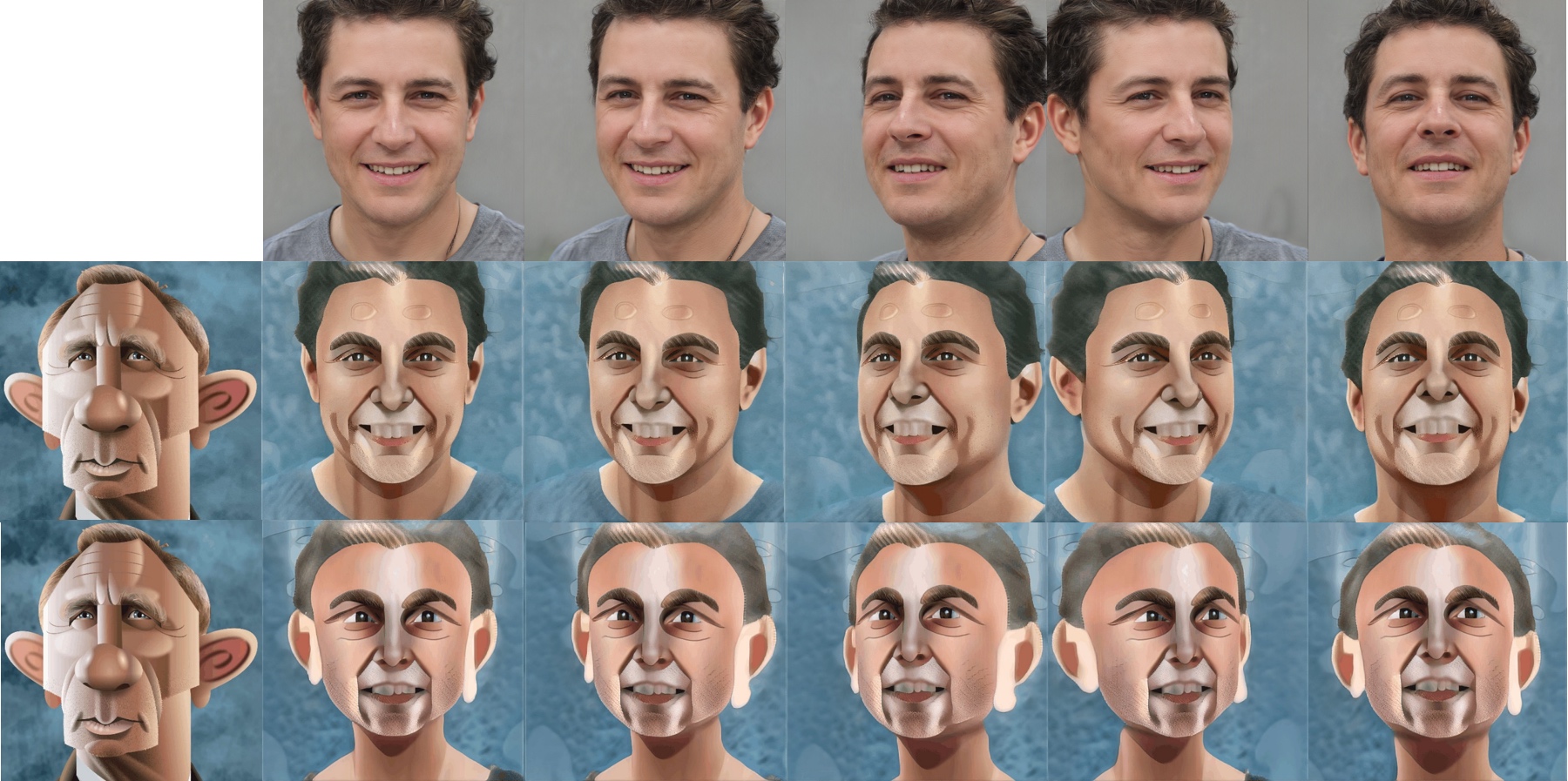}
    \caption{
    \textbf{Validation of geometry regularization.} Geometry change (third row) using the geometry layers discussed in the main paper. Here the style (second row) is changed using the texture layers and the geometry (third row) is changed using the geometry layers. Note that the geometry is not correct as we apply \textit{Style Loss} using a single image and is only used to demonstrate the usage of different layers.
   }
    \label{fig:geo}
\end{figure}

To validate that our chosen layers in Sec.~\ref{sec:loss} are responsible for geometrical and texture changes, we resort to a stylization technique. For stylization given an arbitrary reference image \eg painting, we use the \textit{Style Loss}~\cite{gatys2016image} to update the layers of $G_s$ or $G_t$. Essentially, we use the same parameters used in Sec.~\ref{sec:camera}.  A critical technique to achieve multi-view consistency and circumvent the ghosting face artifact due to single image overfitting is to rotate the camera to cover the $\theta'$ and $\phi'$ ranges in Sec.~\ref{sec:camera} in the main paper uniformly during the optimization. In Fig.~\ref{fig:texture}, we show some results using only the layers used in \textit{Texture Regularization} (Sec.~\ref{sec:loss}). Note the high-quality texture change in the images. In Fig.~\ref{fig:geo}, we show results by adding layers of \textit{Geometry Regularization} (Sec.~\ref{sec:loss}). Note that the geometry is changed in the examples when we use this module. Note that in this example, the geometry is not expected to match as there is no such loss in the optimization. This example results in some arbitrary geometry change that is not flat. This validates our choice of geometry and texture layers used in this paper.

\input{tables/2D-comp}

\section{Failure cases}

\begin{figure}[t]
    \centering
    \includegraphics[width=\linewidth]{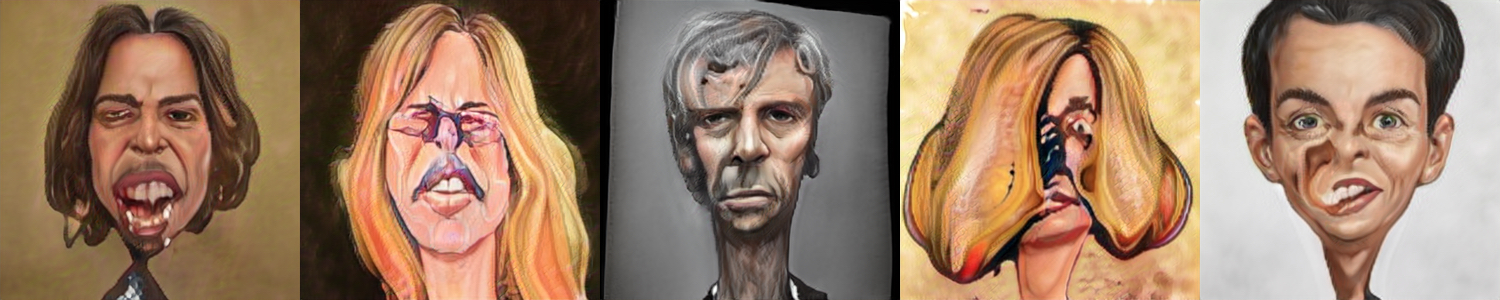}
    \caption{
    \textbf{Failure cases.} Failure cases in caricature generation that stems from the artifacts in the 2D-GAN from which the dataset is generated. Here the artifacts are the result of the generated results of StyleCariGAN~\cite{Jang2021StyleCari} 
   }
    \label{fig:failure}
\end{figure}

Our method has some failure cases that stem from the samples of the 2D-GANs \ie DualStyleGAN~\cite{yang2022Pastiche} and StyleCariGAN~\cite{Jang2021StyleCari}. In the caricature domain, the random samples generated in the case of $\mathrm{G_{t}}$ trained on StyleCariGAN samples can have some severe artifacts (See Fig.~\ref{fig:failure}). Although these do not appear often, such a sample can be improved by attribute classifier and depth regularization losses used on a single image as discussed in Sec. 4 of the main paper. 

\begin{figure*}[t]
    \centering
    \includegraphics[width=0.85\linewidth]{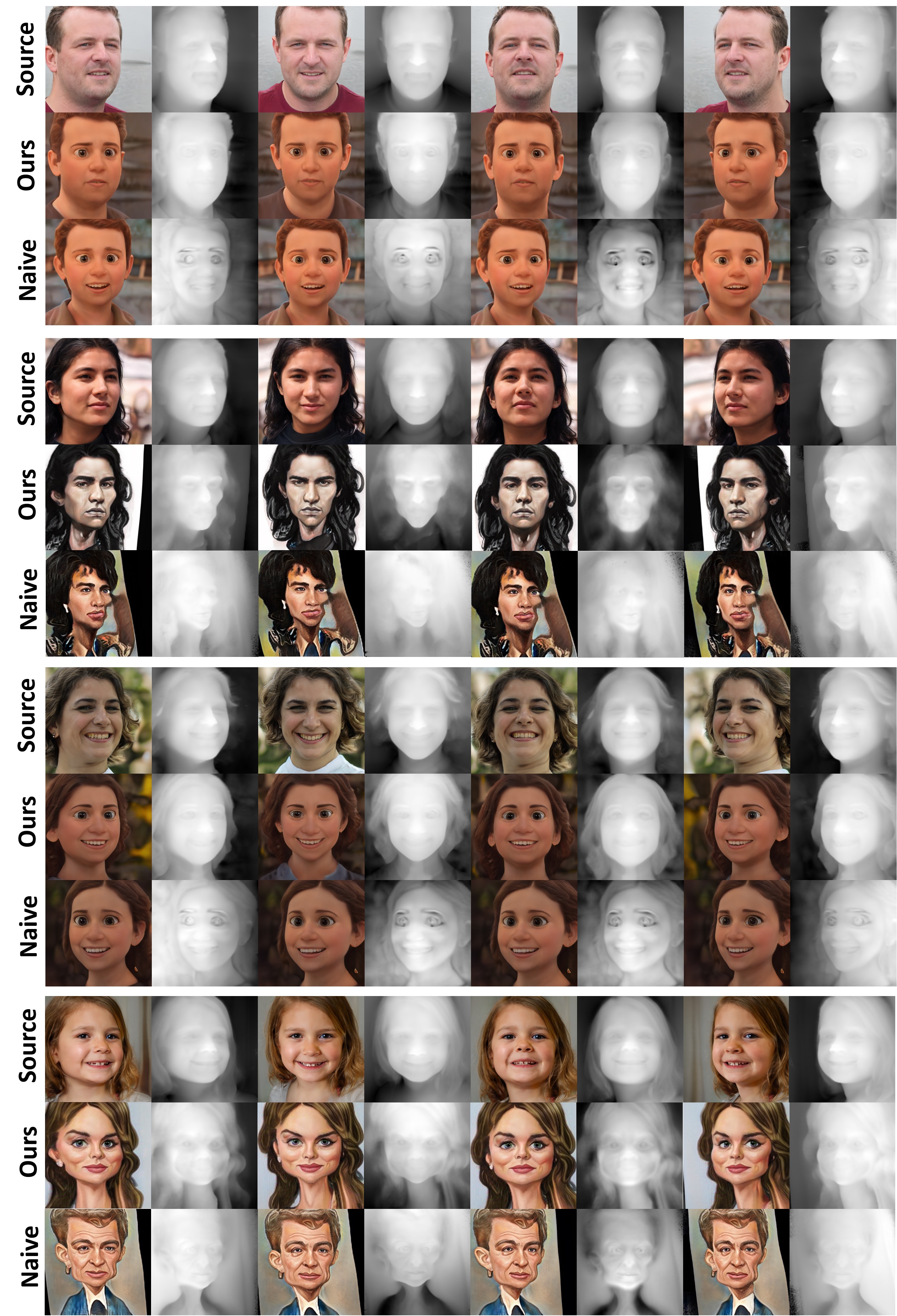}
    \caption{
    \textbf{Comparison with Naive method.} Results of the Caricatures and Pixar toons were viewed from different viewpoints and compared with the baseline method. Note that the depth maps are also visualized highlighting flat geometry. For more results refer to the videos on the \href{https:/rameenabdal.github.io/3DAvatarGAN}{Project Page}.   
   }
    \label{fig:depth}
\end{figure*}

 \section{Depth Map visualization}

\begin{figure*}[t]
    \centering
    \includegraphics[width=\linewidth]{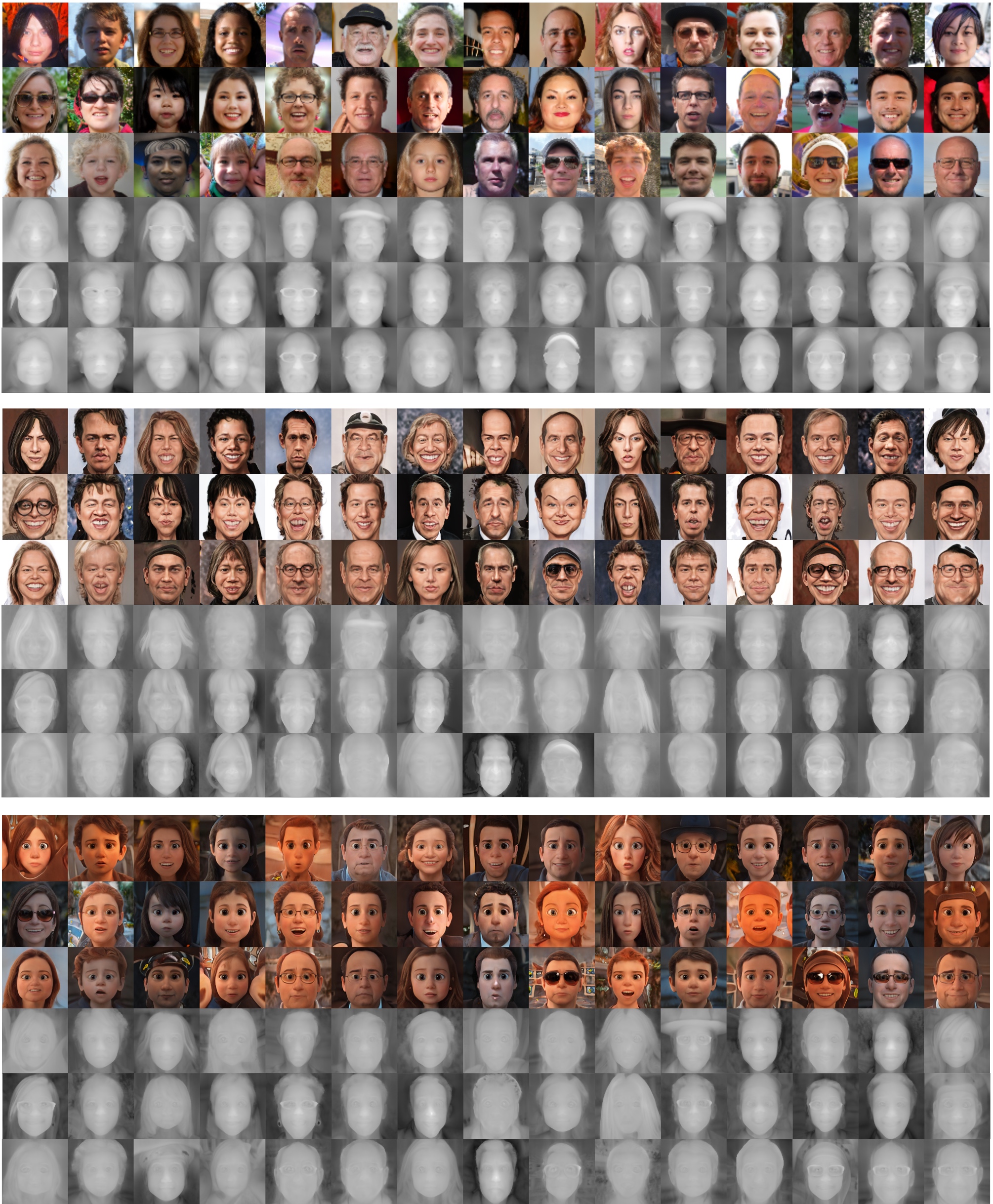}
    \caption{
    \textbf{Grid samples.} Samples from the source domain and corresponding results in the target domain. Corresponding images and depth outputs of the Caricatures and Pixar Toons are shown. 
   }
    \label{fig:sample1}
\end{figure*}

\begin{figure*}[t]
    \centering
    \includegraphics[width=\linewidth]{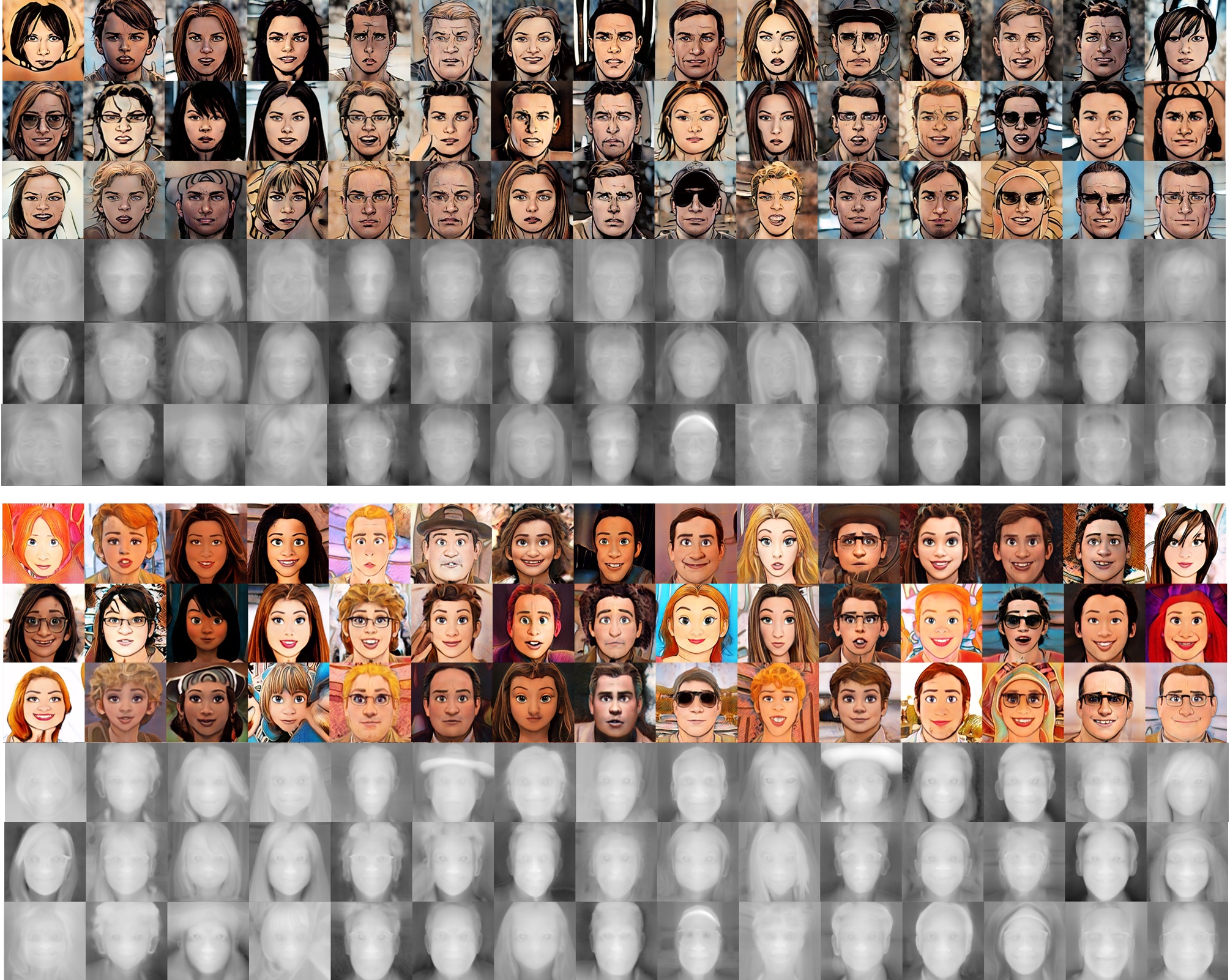}
    \caption{
\textbf{Grid samples.} Extension of Fig. 18. Corresponding images and depth outputs of the Comics and Cartoons are shown. }
    \label{fig:sample2}
\end{figure*}

 In order to show some more samples and the corresponding depth maps for $\mathrm{G_{base}}$ and $\mathrm{G_{t}}$, in Fig.~\ref{fig:depth}, we show some samples from both the generators and the corresponding depth maps with pose changes. Notice the flat geometry in the case of $\mathrm{G_{base}}$ results. Next, in Fig.~\ref{fig:sample1} and Fig.~\ref{fig:sample2}, we show some grid samples of our method with depth maps on the Caricature, Pixar toon, Cartoon, and Comic datasets.

\section{Video Results}
We also show our 3D avatar editing results in videos. We design a simple UI to show that the avatars can be edited in an interactive manner. Please refer to the \href{https:/rameenabdal.github.io/3DAvatarGAN}{webpage} for editing videos and the interactive editing sessions.

%% file: tables/2D-comp.tex
\vspace*{-7mm}
\begin{table}%
\scriptsize
\caption{ 
\textbf{FID comparison with SCG: StyleCariGAN, DSG: DualStyleGAN.}}
\vspace*{-3mm}
\label{tab:2d_comp}
\begin{minipage}{\columnwidth}
\begin{center}
\begin{tabular}{rllll}

  \toprule
  Method  & Cari. (SCG) & Cari. (DSG)      &  Pixar (DSG) &  Cartoons (DSG)\\ \midrule
2D-GANs  & \textbf{51.68} & 96.49   &166.78 & \textbf{105.57} \\
Ours   &  62.90 & \textbf{89.73} & \textbf{162.06} & 111.35 \\

  \bottomrule
\end{tabular} 
\end{center}
\end{minipage}
\end{table}%